\documentclass[10pt,twocolumn,letterpaper]{article}

\usepackage{templates/wacv}
\usepackage{times}
\usepackage{epsfig}
\usepackage{graphicx}
\usepackage{amsmath}
\usepackage{amssymb}
\usepackage{color}
\usepackage{multirow}
\usepackage{hhline}
\usepackage{subcaption}
\newcommand{\rulesep}{\unskip\ \vrule\ }
\usepackage{tablefootnote}
\usepackage{placeins}

\usepackage[pagebackref=true,breaklinks=true,letterpaper=true,colorlinks,bookmarks=false]{hyperref}

\wacvfinalcopy 

\def\wacvPaperID{***} 

\ifwacvfinal\fi
\begin{document}

\title{StuffNet: Using `Stuff' to Improve Object Detection}

\author{Samarth Brahmbhatt\\
Georgia Institute of Technology\\
{\tt\small samarth.robo@gatech.edu}
\and
Henrik I. Christensen\\
UC San Diego\\
{\tt\small hichristensen@ucsd.edu}
\and
James Hays\\
Georgia Institute of Technology\\
{\tt\small hays@gatech.edu}
}

\maketitle
\ifwacvfinal\thispagestyle{empty}\fi

\begin{abstract}
We propose a Convolutional Neural Network (CNN) based algorithm -- StuffNet -- for object detection. In addition to the standard convolutional features trained for region proposal and object detection~\cite{faster_rcnn}, StuffNet uses convolutional features trained for segmentation of objects and `stuff' (amorphous categories such as ground and water). Through experiments on Pascal VOC 2010, we show the importance of features learnt from stuff segmentation for improving object detection performance. StuffNet improves performance from 18.8\% mAP to 23.9\% mAP for small objects. We also devise a method to train StuffNet on datasets that do not have stuff segmentation labels. Through experiments on Pascal VOC 2007 and 2012, we demonstrate the effectiveness of this method and show that StuffNet also significantly improves object detection performance on such datasets.
\end{abstract}

\section{Introduction} \label{sec:1}
\begin{figure}
	\centering
	\begin{subfigure}[b]{0.30\textwidth}
		\includegraphics[width=\textwidth]{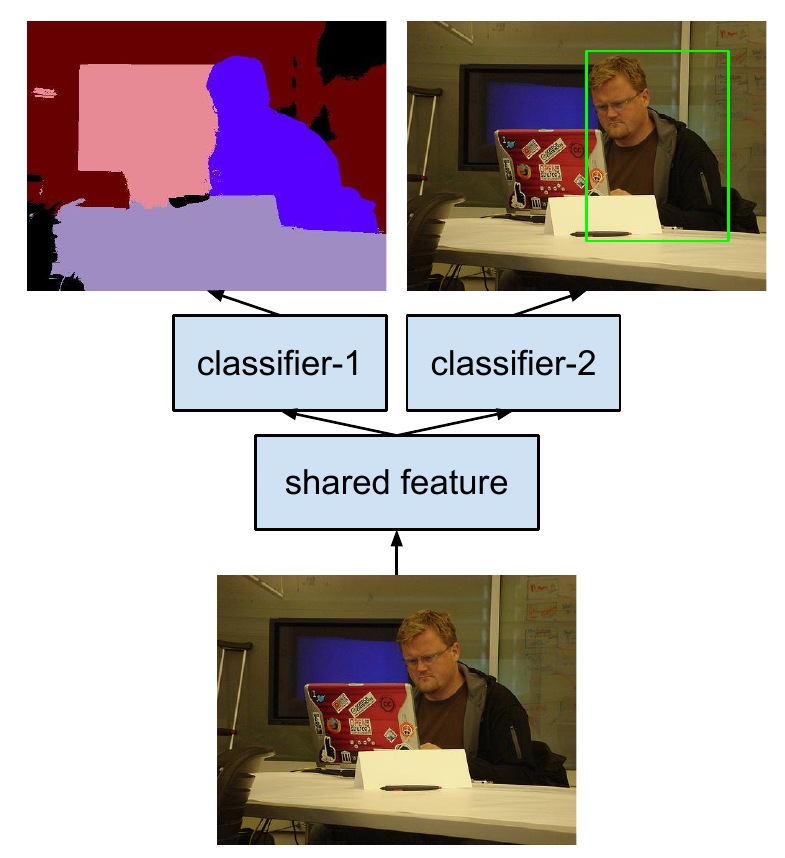}
	\end{subfigure}
	\begin{subfigure}[b]{0.16\textwidth}
		\includegraphics[width=\textwidth]{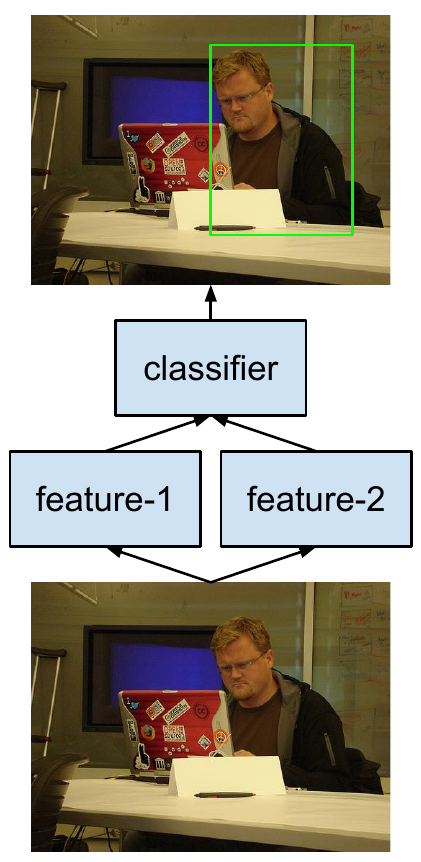}
	\end{subfigure}
\caption{Multi-task learning (left) and multi-feature classification (right)}
\label{fig:2}
\end{figure}

Recent advances in CNN-based object detection have resulted in significant accuracy increases. Faster R-CNN with its integrated region proposal network and end-to-end training scheme for both the object detection and region proposal networks sets an impressive baseline for many object detection datasets. In search for further improvement, the community has mainly chosen three paths: 1) \textit{Deeper networks}: Most notably, ResNet~\cite{resnet} uses 101 layer deep CNNs trained using residual learning to learn more complex representations from images, 2) \textit{Multi-task learning and multi-feature classification} (Figure~\ref{fig:2}):  Multi-task learning refers to using different tasks to provide extra supervision and regularization for object detection. This is done by training the network for an auxiliary task in addition to object detection. The most notable examples of these auxiliary tasks are region proposal~\cite{fast_rcnn, faster_rcnn}, object semantic segmentation~\cite{komodakis_context}, object attribute prediction~\cite{farhadi_attributes, cross_stitch} and face landmark prediction~\cite{face1, face2} for face detection. Multi-feature classification refers to combining diverse features with features trained for object detection at the input of the classifier. These features capture complementary information like context~\cite{segdeepm}, semantic segmentation of surrounding~\cite{mottaghi} and of objects~\cite{komodakis_context}. Multi-task learning algorithms have multiple supervised outputs and typically increase performance for all tasks by learning shared features. In contrast, multi-feature learning algorithms have multiple complementary intermediate features, but don't necessarily have extra supervision. 3) \textit{Post-processing}: These approaches mainly try to better localize the objects by developing iterative localization algorithms that operate on object detections predicted by a base network and often achieve large performance gains~\cite{komodakis_context}.\par
\begin{figure}[t!]
	\centering
	\begin{subfigure}[b]{0.24\textwidth}
		\includegraphics[width=\textwidth]{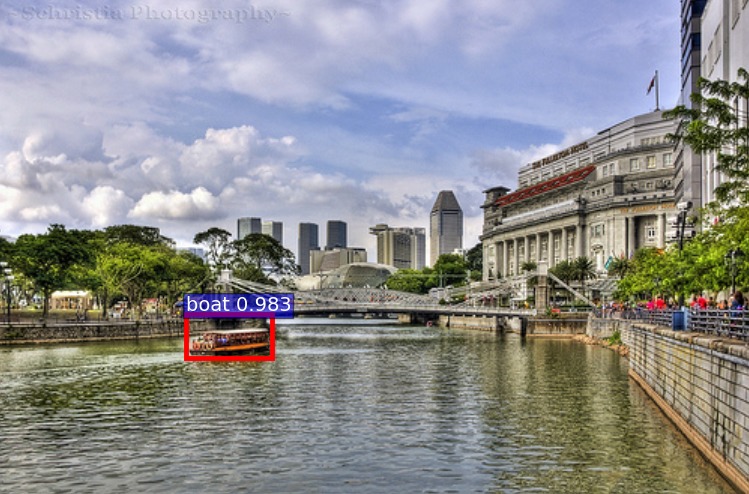}
	\end{subfigure}
	\begin{subfigure}[b]{0.23\textwidth}
		\includegraphics[width=\textwidth]{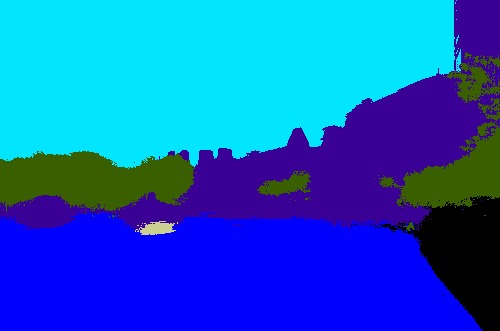}
	\end{subfigure}
\caption{Most objects in images are surrounded by `stuff' (e.g. the boat here is surrounded by water and building). We propose to allow stuff to influence the object detection decision. The figure shows the detection and semantic segmentation output of our model \texttt{StuffNet-30} trained on Pascal VOC 2010 \texttt{trainval} on an image from MS COCO. See Figure~\ref{fig:5e} for legend.}
\label{fig:1}
\end{figure}
This paper combines multi-task learning and multi-feature classification through a task which has been previously omitted from CNN-based object detection frameworks: segmentation of `stuff'. Stuff~\cite{stuff} refers to object categories which do not have a fixed shape and hence are difficult to bound with boxes e.g. water, sky, wall etc. We examine this task for two reasons: 1) It has been shown both in computer vision~\cite{stuff, mottaghi, segdeepm, context_survey} and cognitive psychology research~\cite{olivia_context_survey} that identifying the local surrounding of an object helps to better identify it. This is because object identity tends to be correlated with spatially surrounding stuff, e.g. boats in most 2D images have water below them and sky/land above them (Figure~\ref{fig:1}). Especially for small, occluded, or unusually posed objects the local appearance might be ambiguous but context (in the form of surrounding stuff) can clarify what objects are likely. Hence, conditioning an object detection decision on features trained to identify stuff is likely to be beneficial. We demonstrate this through extensive evaluation in Section~\ref{sec:4}. Figure~\ref{fig:4} also shows this qualitatively. 2) There have been many recent advancements in CNN-based semantic segmentation of images~\cite{fcn, deeplab, parsenet, cnn_crf}. Even though they are mainly focused on segmenting the objects, the network architectures and training mechanisms can be used just as easily for semantic segmentation of stuff. 3) Since the appearance of stuff (like sky, water, wall, trees, etc.) is similar across images from different datasets, a network trained to segment stuff using labels from one dataset can generalize well to a different dataset without re-training (we show this qualitatively in Figure~\ref{fig:5}). This means that once the stuff segmentation part of StuffNet is trained, we do not need stuff labels to train StuffNet on a different dataset. This is in contrast with most other tasks used for multi-task learning, and offsets the lack of large scale training data for stuff segmentation. Out of the numerous object detection datasets available to the community (Pascal VOC~\cite{voc-2007, voc-2010, voc-2012}, MS COCO~\cite{mscoco}, ImageNet~\cite{imagenet}, Cityscapes~\cite{cityscapes}, etc.), only Pascal 2010 (through Pascal Context~\cite{mottaghi}) and Cityscapes have stuff annotations. In this paper we train the stuff segmentation part of StuffNet only on the stuff labels in the Pascal Context dataset, and show that this can be deployed to Pascal VOC 2007 and Pascal VOC 2012 to boost object detection performance.\par
To summarize, we make the following contributions in this paper:
\begin{itemize}\itemsep0em
\item \textit{CNN architecture}: Our CNN architecture (Figure~\ref{fig:3}) supports both multi-task learning and multi-feature classification. We augment Faster R-CNN~\cite{faster_rcnn} with a semantic segmentation branch based on DeepLab~\cite{deeplab}. This creates CNN that can learn both object detection and stuff segmentation in an end-to-end manner, allowing gradients from both tasks to influence a shared feature map (multi-task learning). Faster R-CNN uses convolutional feature maps to propose possible object regions and make decisions about the class of object present in those regions. These convolutional feature maps are trained solely for the tasks of region proposal and classification. In contrast, the StuffNet architecture conditions the object detection additionally on convolutional feature maps trained for stuff segmentation, and optionally, object segmentation (multi-feature classification). Through experiments on various datasets, we demonstrate that multi-task learning and multi-feature classification improves object detection performance.
\item \textit{Effective use of stuff labels}: Availability of stuff segmentation labels is a concern while training on a dataset. However, since stuff appears same across datasets, we only need one dataset with stuff labels to train the stuff segmentation part of StuffNet. We show that while training on a new dataset that does not have stuff labels, constraining the conv feature maps to produce the same segmentation output as a network trained on the earlier dataset acts as an effective regularizer and boosts object detection performance.
\end{itemize}
The rest of the paper is organized as follows: Section~\ref{sec:2} reviews related work in this area. Section~\ref{sec:3} describes our CNN architecture and training procedure. Section~\ref{sec:4} shows results on object detection datasets and Section~\ref{sec:5} discusses our results.
\section{Related Work} \label{sec:2}
\subsection{CNN-based object detection}
Fast R-CNN~\cite{fast_rcnn} pools convolutional feature maps using external region proposals (selective search~\cite{selective_search}), and classifies them using fully connected layers interspersed with non-linearities. Faster R-CNN~\cite{faster_rcnn} incorporates the region proposal process into the network with a region proposal network. In this paper, we use Faster R-CNN as a baseline since it delivers state-of-the-art performance across many datasets (excluding deep residual networks) and has publicly available code. Recent advances with deep residual learning have resulted in 101 layer deep residual networks~\cite{resnet} that deliver superior performance. Even though we do not employ residual learning in this paper, our algorithm can be easily applied to residual nets to harness their deeper representations, since they have been applied to both object detection~\cite{resnet} and semantic segmentation~\cite{resnet_seg}.
\subsection{Multi-task learning for improving object detection}
Multi-task learning is the process of learning a representation that is useful for multiple tasks from the same input and has been found to improve the performance of both tasks~\cite{multi_1, multi_2, multi_3, multi_4}. The most notable recent examples of multi task learning improving object detection are Fast R-CNN~\cite{fast_rcnn} and Faster R-CNN~\cite{faster_rcnn}. Fast R-CNN shows that training simultaneously for object detection and bounding box regression improves object detection performance. Faster R-CNN shows improvements in object detection performance by training for region proposal and bounding box regression. Recently, it has been shown that learning to predict facial landmarks along with learning to detect faces can improve face detection performance~\cite{face1, face2}.
\subsection{Multi-feature classification for improving object detection}
Recent works have shown that using features targeted at different tasks as input to the final classification layers of an object detector improves performance. For example,~\cite{mottaghi} uses counts of pixels around the object that are labeled with different stuff categories.~\cite{segdeepm} uses features from the penultimate layer of AlexNet extracted from a ring shaped region around the object.~\cite{komodakis_context} uses convolutional features from a network trained separately for object segmentation, extracted from an expanded region proposal around the object. We compare the performance of StuffNet to~\cite{segdeepm} and~\cite{komodakis_context} and show that StuffNet performs better.
\section{StuffNet Architecture and Training} \label{sec:3}
\begin{figure*}
  \centering
  \begin{subfigure}[b]{0.49\textwidth}
		\includegraphics[width=\textwidth]{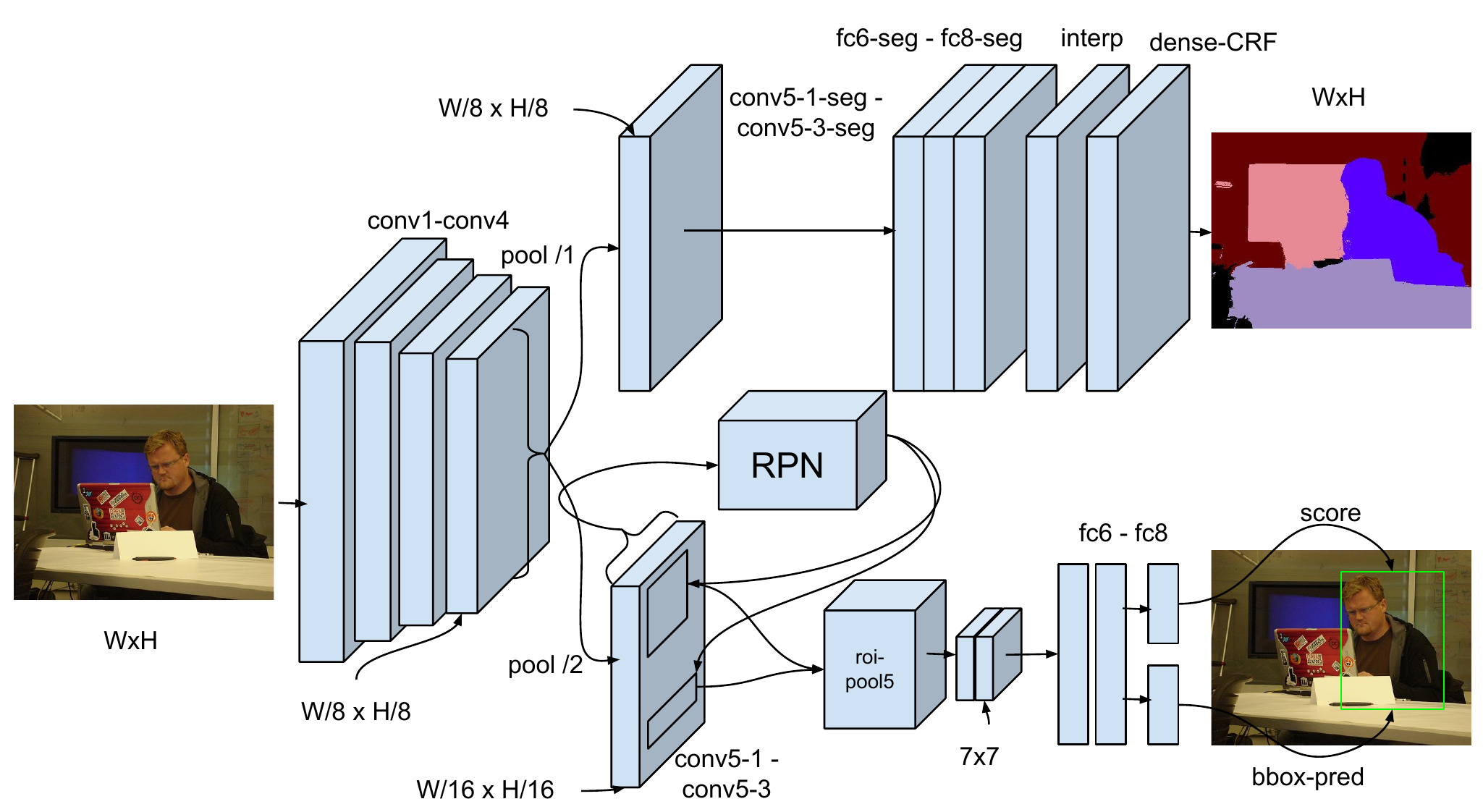}
		\caption{\texttt{StuffNet-multitask}}
		\label{fig:3a}
	\end{subfigure}
	\rulesep
	\begin{subfigure}[b]{0.49\textwidth}
		\includegraphics[width=\textwidth]{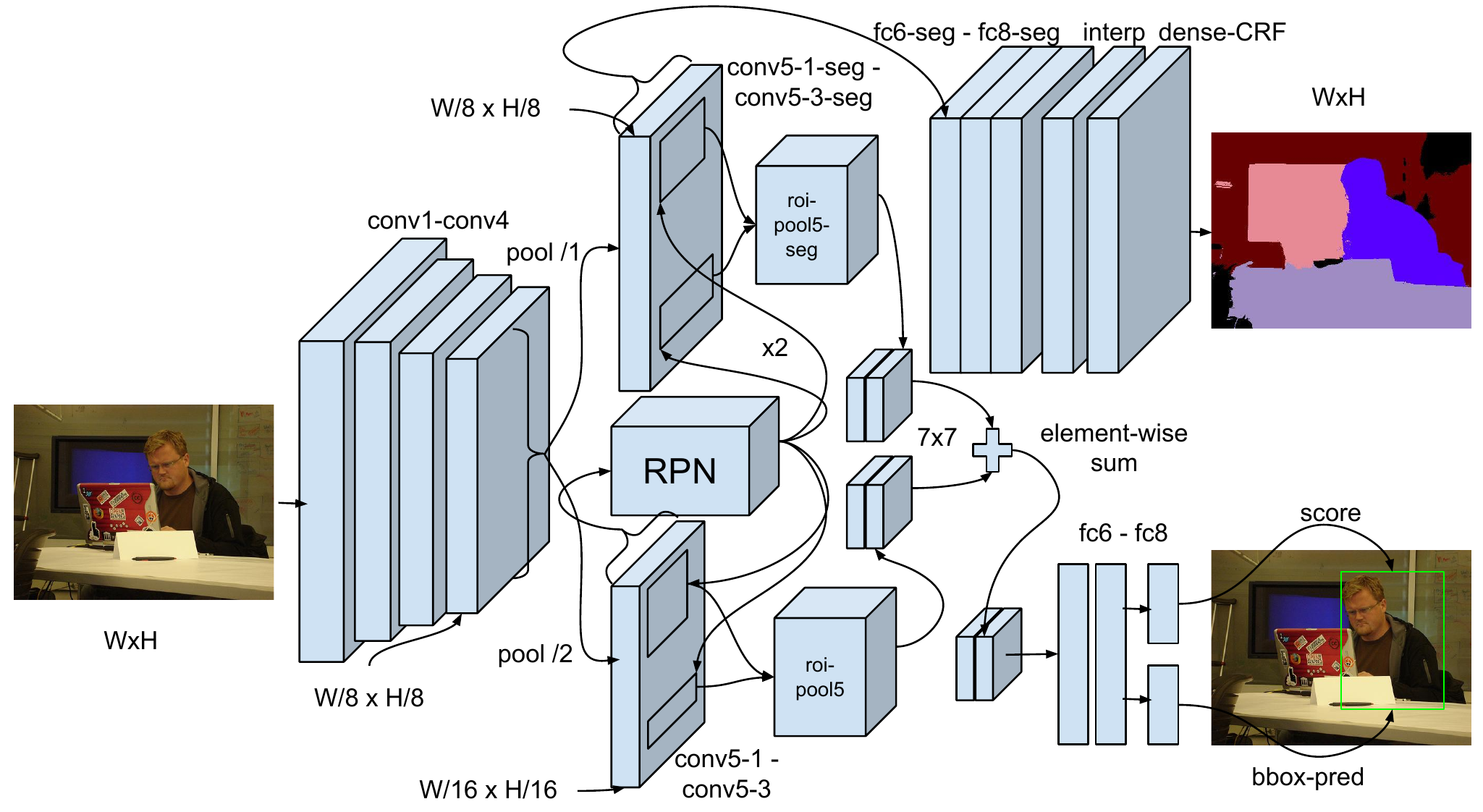}
		\caption{\texttt{StuffNet-10}}
		\label{fig:3b}
	\end{subfigure}
\caption{CNN architecture for two variants of \texttt{StuffNet}. \texttt{StuffNet-multitask} implements multi-task learning. \texttt{StuffNet-10} implements both multi-task learning and use of diverse features. Please see~\cite{faster_rcnn} for details on the Region Proposal Network (RPN) and~\cite{deeplab} for details on the dense CRF.}
\label{fig:3}
\end{figure*}
Stuffnet is a two-branch network which simultaneously performs object detection and semantic segmentation. The object detection branch is based on Faster R-CNN~\cite{faster_rcnn} and the semantic segmentation branch is based on DeepLab~\cite{deeplab}. We first explain them briefly.
\subsection{Faster R-CNN}\label{sec:3-1}
\textbf{Why Faster R-CNN?} Faster R-CNN is the state of the art in object detection and integrates the region proposal and region classification tasks, which we traditionally performed separately. Most of the entries winning the MS COCO object detection challenge are based on Faster R-CNN, including~\cite{resnet}, which uses a deep residual network to learn features but otherwise uses the same workflow. Hence Faster R-CNN makes a strong baseline to compare against. Moreover, the region proposal and classification modules of Faster R-CNN use the same input features. By training these features for diverse tasks (as described in Section~\ref{sec:3-3}), it is possible to influence both the region proposal and classification processes.
\subsubsection{Faster R-CNN architecture}\label{sec:3-1-2}
The network consists of a Region Proposal Network (RPN)~\cite{faster_rcnn} which proposes bounding boxes likely to contain an object, and the Fast R-CNN~\cite{fast_rcnn} detector that classifies the proposals (see bottom branch of Figure~\ref{fig:3a}). Both the RPN and Fast R-CNN share the \texttt{conv5-3} feature map from the VGG 16-layer CNN~\cite{vgg} as input.\par
The RPN is a fully convolutional network that produces a set of bounding boxes and their objectness scores. It slides a 3x3 window over the \texttt{conv5-3} feature map and maps it from 1024 channels to 512 channels using a 3x3 conv layer. Next, two 1x1 conv layers are used in parallel. The \texttt{rpn-cls} layer predicts $2k$ object presence/absence scores and the \texttt{rpn-reg} layer predicts $4k$ co-ordinate regression parameters for $k$ anchor proposal boxes. Each anchor is centered at the corresponding 3x3 window, and has a unique scale and aspect ratio. The $4k$ co-ordinate regression parameters are applied to the $k$ anchors to produce the final region proposals. The $2k$ object presence/absence scores are passed through a softmax layer to get the objectness scores.\par
The Fast R-CNN detector places an ROI pooling layer \texttt{roi-pool5}~\cite{sppnet, fast_rcnn} on top of \texttt{conv5-3}. This ROI pooling layer also takes the minibatch of region proposals as input. It adaptively max-pools the area of the activation map under each proposal into a grid equal to the input size of the following fully connected layer (7x7 for VGG-16 \texttt{fc6}). The final fully connected layer gives the object classification scores. Rectified activations of the \texttt{fc7} units are also passed to a bounding box regression layer \texttt{bbox-pred}, which predicts parameters to change the four corners of the object proposal, separately for each class.\par
Faster R-CNN minimizes a weighted sum of four loss functions: 1) softmax loss for presence/absence of a foreground object in the region proposals, 2) smooth L1 loss~\cite{fast_rcnn} for anchor co-ordinate regression, 3) softmax loss for the classification of region proposals and 4) smooth L1 loss for region proposal co-ordinate regression. 
\subsubsection{Faster R-CNN training}
We use the publicly available Python implementation of Faster R-CNN, which implements \textit{joint approximate training}. The RPN training minibatch consists of 256 proposals from one image with a 1:1 ratio of positives (IoU with a ground-truth box $>$ 0.7) to negatives (IoU $<$ 0.3). The Fast R-CNN training minibatch has 128 proposals with a 1:3 ratio of foreground (IoU with a ground-truth box of the same class $>$ 0.5) to background (0.1 $<$ IoU with any foreground ground-truth box $<$ 0.5). For more more details on training parameters, please see~\cite{faster_rcnn}. 
\subsection{DeepLab}\label{sec:3-2}
\textbf{Why DeepLab?} DeepLab~\cite{deeplab} provides a fully convolutional~\cite{fcn} architecture based on the VGG 16-layer CNN that has been shown to produce state-of-the-art object segmentation results. It can be easily modified to train for stuff semantic segmentation by providing stuff segmentation labels for training and changing the channel dimension of the last 1x1 convolution layer.
\subsubsection{DeepLab architecture}\label{sec:3-2-2}
In this paper, we use the \texttt{Deeplab-LargeFOV}~\cite{deeplab} architecture (shown in the top branch of Figure~\ref{fig:3a}). DeepLab `convolutionizes' the VGG 16-layer architecture~\cite{vgg} by using convolutions instead of fully connected layers \texttt{fc6} - \texttt{fc8}. However, since spatial subsampling in max-pooling layers reduces the resolution of the feature map input to \texttt{fc6} by a factor of 32 for the VGG architecture, this leads to a very coarse segmentation output. DeepLab solves this problem by skipping spatial subsampling in max-pooling after the \texttt{conv4-3} and \texttt{conv5-3} layers and introducing `holes'~\cite{holes,deeplab} in the convolutional filters of layers downstream from \texttt{conv4-3}. Thus the resolution of the score map output by the DeepLab \texttt{fc8} layer is reduced by only a factor of 8. This score map is up-sampled using bilinear interpolation to get the final stuff segmentation output. At test time, DeepLab also applies a dense CRF~\cite{dense_crf} on it's output to get crisp object boundaries. Finally, DeepLab minimizes a softmax loss over the segmentation label of each pixel in the upsampled score map.
\subsubsection{DeepLab training}
DeepLab uses a simple training procedure that involves separate training for the CNN and the dense CRF. The CNN is trained using stochastic gradient descent with a minibatch of 10 images to provide unaries for the CNN. Subsequently, the parameters of the dense CRF are picked using cross validation, assuming the unary terms provided by the CNN are fixed. For more details, please see~\cite{deeplab}.
\subsection{StuffNet}\label{sec:3-3}
In this section we describe the changes we make to the architecture and training procedures of Faster R-CNN and DeepLab to construct StuffNet and train it in an end-to-end fashion.
\subsubsection{StuffNet architecture}\label{sec:3-3-1}
The objective of StuffNet is to improve object detection by utilizing 1) multi-task learning and 2) multi-feature classification. Towards this end, we first propose a simple combination of Faster R-CNN and DeepLab that implements multi-task learning, which we call \texttt{StuffNet-multitask}. Next we build on top of \texttt{StuffNet-multitask} and propose an architecture that incorporates semantic segmentation features into the detection decision process, which we call \texttt{StuffNet-10} and \texttt{StuffNet-30}.\par
\textbf{Multi-task Learning.} As shown in Figure~\ref{fig:3a}, we combine Faster R-CNN and DeepLab to create \texttt{StuffNet-multitask}. Since both networks are based on the VGG 16-layer architecture, they share the conv layers up to \texttt{conv4-3}. This allows StuffNet-multitask to learn shared features that are useful for both object detection and semantic segmentation. Next, we pool the \texttt{conv4-3} feature map twice: once with a stride of 2 units (\texttt{pool4-3}), and once with a stride of 1 unit (\texttt{pool4-3-seg}) to produce the inputs to the next conv layers of Faster R-CNN and DeepLab respectively. This is essential to avoid a very coarse resolution at the DeepLab output, as described in Section~\ref{sec:3-2-2}. The downstream layers of Faster R-CNN and DeepLab follow their respective architectures as described in Sections~\ref{sec:3-1-2} and~\ref{sec:3-2-2} respectively. Names of layers in the DeepLab branch have a \texttt{-seg} appended to distinguish them from corresponding layers in the Faster R-CNN branch. \texttt{StuffNet-multitask} minimizes the sum of the Faster R-CNN and DeepLab loss functions.\par
\textbf{Multi-feature classification.} \texttt{StuffNet}-\texttt{multitask} allows to learn a shared convolutional feature map that is used by both the object detection and semantic segmentation networks. However, it does not allow segmentation to directly influence the detection decision. We allow this influence in \texttt{StuffNet-10} and \texttt{StuffNet-30} by creating a connection between the two branches of StuffNet before the \texttt{fc6} layer, as shown in Figure~\ref{fig:3b}. We introduce another roi-pooling layer \texttt{roi-pool5-seg} that adaptively max-pools features from the \texttt{conv5-3-seg} layer. The regions input to \texttt{roi-pool5-seg} are double the size of those input to \texttt{roi-pool5}, to account for the higher resolution of the \texttt{conv5-3-seg} feature map. However, the output size of this roi-pooling layer is also set to 7x7, which allows us to combine the outputs of both roi-pooling layers by an \textit{element-wise sum}. This combined representation is input to the \texttt{fc6} layer of the Faster R-CNN branch. We tried a non-linear combination of features as mentioned in~\cite{malik_small_context}, but found that a linear combination gives better performance and allows faster learning.\par
\textbf{Classes for semantic segmentation.} The main focus of this paper is to show that identifying the stuff around objects helps better identify them. Indeed,~\cite{mottaghi} have shown this using hand-crafted features. However, recently, some CNN-based object detection networks~\cite{komodakis_context} have shown performance improvements by including features trained for the segmenting the same objects that are being detected. Hence we develop two variants of StuffNet: \texttt{StuffNet-10} segments the image into 10 stuff classes, while \texttt{StuffNet-30} segments the image into 10 stuff classes + 20 Pascal VOC object classes. We identify the 10 stuff classes from the main 33 classes used in the Pascal Context dataset~\cite{mottaghi}. They are: \textit{wall, floor, water, tree, sky, road, ground, building, mountain} and \textit{background}. Since the annotators for the Pascal Context dataset were free to create their own label names, we merge some categories into the 9 stuff categories mentioned above. Specifically, we merge \textit{sidewalk} and \textit{runway} into \textit{road}, \textit{ceiling} into \textit{wall}, and \textit{grass, platform, sand, snow} into \textit{ground}.
\subsubsection{StuffNet training}\label{sec:3-3-2}
The learning rate policy, number of iterations and batch size differ greatly for Faster R-CNN~\cite{faster_rcnn} and DeepLab~\cite{deeplab}. Faster R-CNN uses a batch size of 1 image (256 regions for RPN, 128 regions for classification) and trains for 70k iterations. The learning rate starts at 1e-3 and follows `step' learning rate policy, reducing by a factor of 10 after 50k iterations. On the other hand, DeepLab uses a batch size of 10 images, 20k iterations and a `poly' learning rate with the power parameter set to 0.9 (see~\cite{deeplab, parsenet} for details on the `poly' learning rate policy). Since our main objective in this paper is to improve object detection performance, we use all the learning parameters from Faster R-CNN for all variants of StuffNet.\par
During training, StuffNet requires object bounding box labels as well as pixel-level semantic segmentation labels for all the segmentation classes (10 and 30 for \texttt{StuffNet-10} and \texttt{StuffNet-30} respectively). Hence, we first train StuffNet-multitask, StuffNet-10 and StuffNet-30 on the Pascal VOC 2010 dataset\cite{voc-2010}, which has object bounding box labels from the VOC object detection task, object segmentation labels from the VOC object segmentation task, and stuff labels from the Pascal Context dataset~\cite{mottaghi}. This network can then be used to generate `ground-truth' stuff labels for datasets that lack them, as described next.\par
\begin{figure*}[h!]
  \centering
	\begin{subfigure}[b]{0.22\textwidth}
    	\includegraphics[width=\textwidth]{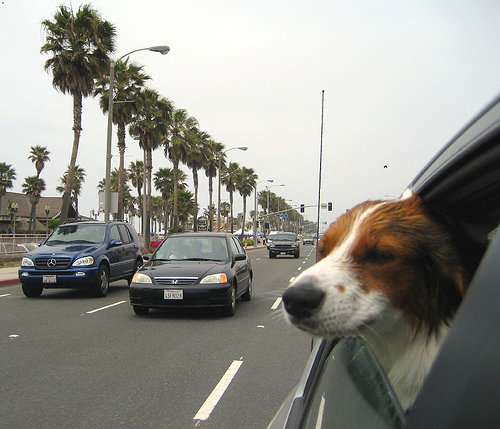}
        \includegraphics[width=\textwidth]{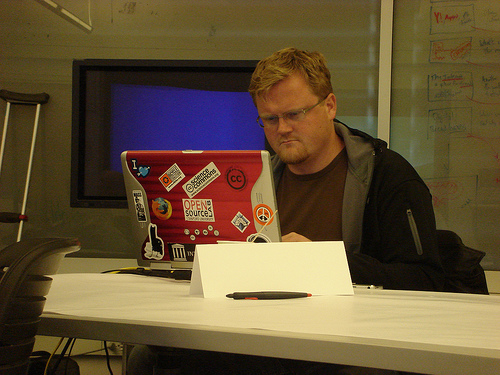}
        \includegraphics[width=\textwidth]{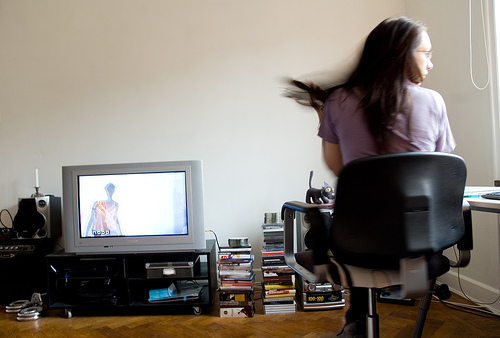}
        \caption{Image}
	\end{subfigure}
	\begin{subfigure}[b]{0.22\textwidth}
		\includegraphics[width=\textwidth]{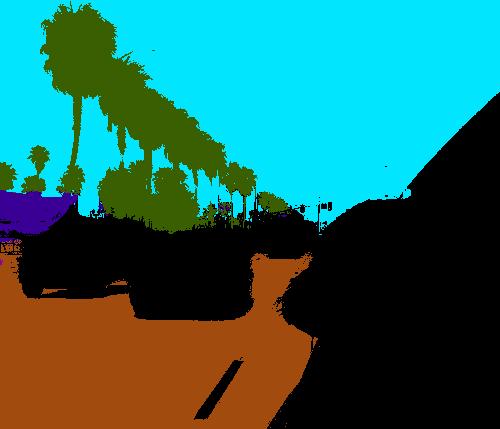}
        \includegraphics[width=\textwidth]{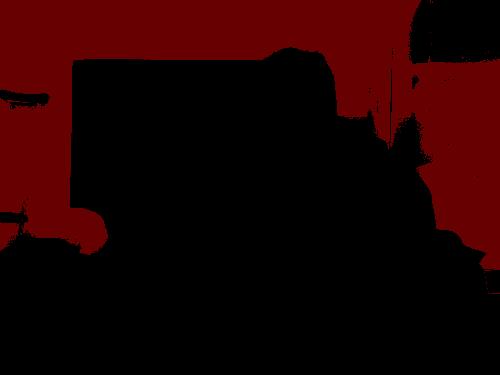}
        \includegraphics[width=\textwidth]{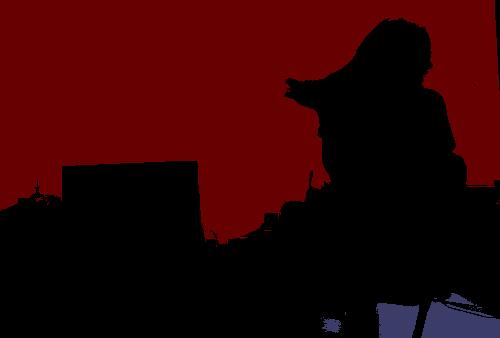}
        \caption{StuffNet-10}
	\end{subfigure}
	\begin{subfigure}[b]{0.22\textwidth}
		\includegraphics[width=\textwidth]{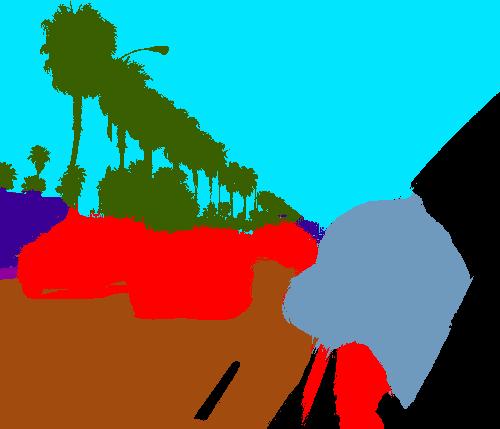}
        \includegraphics[width=\textwidth]{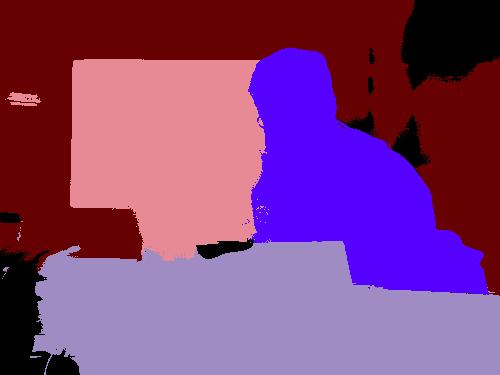}
        \includegraphics[width=\textwidth]{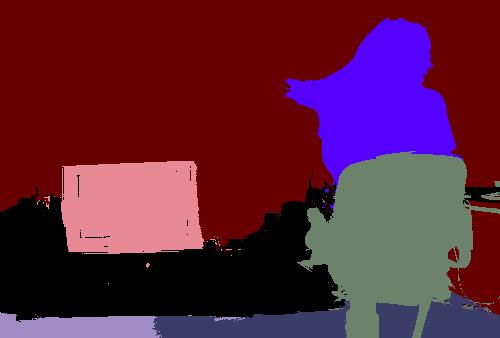}
        \caption{StuffNet-30}
        \label{fig:5c}
	\end{subfigure}
	\begin{subfigure}[b]{0.22\textwidth}
		\includegraphics[width=\textwidth]{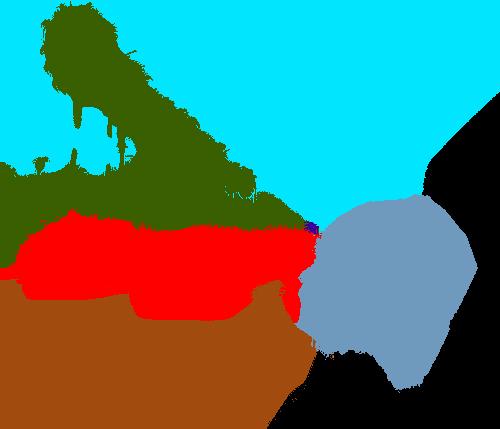}
        \includegraphics[width=\textwidth]{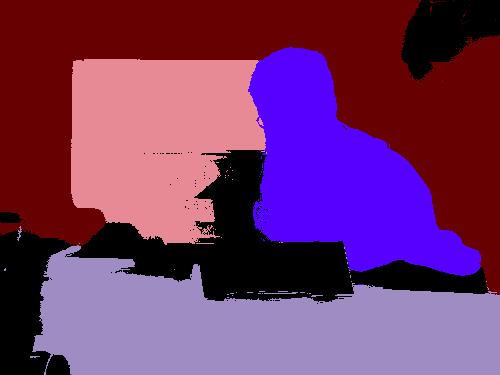}
        \includegraphics[width=\textwidth]{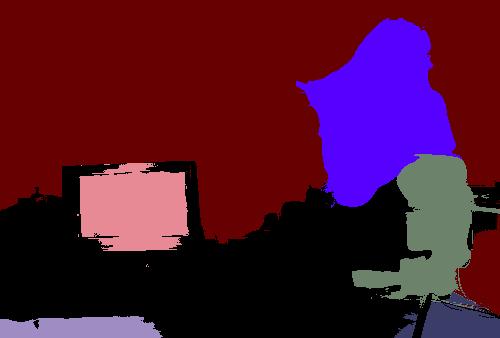}
        \caption{StuffNet-30}
        \label{fig:5d}
	\end{subfigure}
    \begin{subfigure}[b]{0.10\textwidth}
    	\includegraphics[width=\textwidth, height=0.395\textheight]{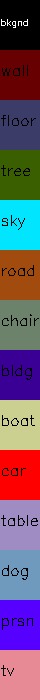}
        \caption{legend}
        \label{fig:5e}
     \end{subfigure}
\caption{StuffNet segmentation qualitative results on images from the MS COCO dataset. Models (b) and (c) were trained on Pascal VOC 2010 \texttt{trainval}, while (d) was trained on Pascal VOC 2012 \texttt{trainval} using feature constraining.}
\label{fig:5}
\end{figure*}

\textbf{Feature Constraining to train StuffNet without segmentation labels.} We use the fact that stuff appears same across datasets to develop a simple procedure to train StuffNet on datasets that lack stuff segmentation labels. We use the DeepLab branch of StuffNet-10 trained on Pascal VOC 2010 to get \textit{hallucinated} stuff ground truth labels for all images in that dataset. As shown qualitatively in Figure~\ref{fig:5}, the stuff labels produced in this manner are very close to ground truth, especially after the use of the dense CRF at the end of DeepLab to capture detailed region boundaries. We then train StuffNet as described previously, using these stuff labels. This constrains the conv layers shared between DeepLab and Faster R-CNN, and the conv layers in the DeepLab branch, to continue producing the same stuff labels as the training iterations proceed. We call this \textit{feature constraining}. Figures~\ref{fig:5c} and~\ref{fig:5d} show the stuff labels produced by StuffNet-10 on an image from MS COCO before and after this training respectively. It can be seen that this simple constraining strategy is effective in practice. An alternative to this method is to freeze the segmentation branch of the network and only re-train the detection branch. However, as shown in the multi-task setup of Figure~\ref{fig:2}, both branches operate on a shared representation. As this shared representation gets updated by the gradients from the detection branch, it no longer remains compatible with the frozen segmentation branch. We develop the feature constraining method to avoid this problem.
\section{Results} \label{sec:4}
\begin{table*}[ht]
\resizebox{\textwidth}{!}{
\begin{tabular}{|c|cccccccccccccccccccc|c|}
\hline
\textbf{Method} & \textbf{areo} & \textbf{bike} & \textbf{bird} & \textbf{boat} & \textbf{bottle} & \textbf{bus} & \textbf{car} & \textbf{cat} & \textbf{chair} & \textbf{cow} & \textbf{table} & \textbf{dog} & \textbf{horse} & \textbf{mbike} & \textbf{person} & \textbf{plant} & \textbf{sheep} & \textbf{sofa} & \textbf{train} & tv & \textbf{mAP}\\
\hline
Faster R-CNN~\cite{faster_rcnn} & 79.9 & 78.4 & 66.6 & 44.7 & \textbf{50.2} & 77.9 & 70.3 & 86.0 & 40.6 & 60.4 & 50.1 & \textbf{83.9} & 76.7 & \textbf{78.8} & 77.4 & 36.0 & 68.8 & 55.4 & 75.6 & 64.5 & 66.1\\
\hline
StuffNet multitask & \textbf{80.3} & 76.5 & 65.0 & 50.3 & 48.4 & \textbf{78.3} & 70.8 & 84.6 & 42.0 & 61.4 & 49.1 & 83.0 & 75.2 & 77.8 & 77.0 & 40.0 & 70.2 & 58.4 & \textbf{79.5} & 64.9 & 66.6\\
StuffNet-10 & 79.8 & 77.8 & \textbf{67.7} & 51.7 & 49.5 & 76.7 & \textbf{72.1} & 84.2 & 44.1 & \textbf{64.5} & 52.1 & 82.6 & 75.8 & 78.4 & 78.0 & \textbf{42.2} & 69.7 & \textbf{58.9} & 78.6 & 64.8 & 67.5\\
StuffNet-30 & 80.0 & \textbf{79.5} & 67.3 & \textbf{53.5} & 49.4 & 77.2 & 71.3 & \textbf{86.4} & \textbf{46.4} & 62.8 & \textbf{54.4} & 83.5 & \textbf{78.8} & 78.6 & \textbf{78.5} & 38.1 & \textbf{70.8} & 58.6 & 79.0 & \textbf{65.4} & \textbf{68.0}\\
\hline
\end{tabular}}
\caption{Ablation study analysing effects of multi-task learning and feature diversification. Trained on Pascal VOC 2010 \texttt{train}, tested on Pascal VOC 2010 \texttt{val}}
\label{tab:1}
\end{table*}
\begin{table*}[ht!]
\resizebox{\textwidth}{!}{
\begin{tabular}{|c|c|cccccccccccccccccccc|c|c|}
\hline
\textbf{Method} & \textbf{Size} & \textbf{areo} & \textbf{bike} & \textbf{bird} & \textbf{boat} & \textbf{bottle} & \textbf{bus} & \textbf{car} & \textbf{cat} & \textbf{chair} & \textbf{cow} & \textbf{table} & \textbf{dog} & \textbf{horse} & \textbf{mbike} & \textbf{person} & \textbf{plant} & \textbf{sheep} & \textbf{sofa} & \textbf{train} & \textbf{tv} & \textbf{mAP} & \textbf{mAP change}\\
\hline
Faster R-CNN~\cite{faster_rcnn} & small & 34.0 & \textbf{16.7} & 7.8 & 27.6 & \textbf{15.7} & \textbf{33.3} & 42.8 & 0.0 & 0.0 & 29.6 & 0.0 & 0.0 & 25.0 & \textbf{16.7} & 30.2 & \textbf{19.8} & 43.3 & 0.0 & \textbf{33.3} & 0.0 & 18.8 & 0.0\\
StuffNet-10 & small & \textbf{35.7} & 0.0 & \textbf{16.9} & \textbf{33.0} & 13.4 & \textbf{33.3} & \textbf{47.9} & \textbf{33.3} & \textbf{3.6} & 46.3 & 0.0 & \textbf{33.3} & \textbf{31.3} & \textbf{16.7} & \textbf{35.5} & 19.4 & \textbf{45.3} & 0.0 & \textbf{33.3} & 0.0 & \textbf{23.9} & \textbf{+5.1}\\
StuffNet-30 & small & 28.1 & 0.0 & 15.5 & 26.5 & 13.4 & 0.0 & 46.0 & \textbf{33.3} & \textbf{3.6} & \textbf{61.5} & 0.0 & \textbf{33.3} & 20.8 & 8.3 & 33.2 & 16.4 & 42.7 & 0.0 & \textbf{33.3} & 0.0 & 20.8 & +2.0\\
\hline
Faster R-CNN~\cite{faster_rcnn} & medium & 53.4 & 67.6 & 59.8 & 48.8 & \textbf{50.1} & 52.9 & 70.1 & \textbf{55.5} & 33.8 & 66.1 & 2.4 & \textbf{74.5} & 60.7 & 57.0 & 68.3 & 45.3 & 69.7 & 13.3 & 30.0 & 56.5 & 51.8 & 0.0\\
StuffNet-10 & medium & \textbf{57.0} & 65.1 & 62.6 & 54.3 & 48.8 & 54.4 & 73.3 & 52.2 & 34.5 & \textbf{69.3} & \textbf{3.8} & 68.0 & 59.7 & \textbf{60.4} & \textbf{71.2} & \textbf{50.8} & 71.0 & \textbf{30.0} & 25.2 & 56.3 & 53.4 & +1.6\\
StuffNet-30 & medium & 56.8 & \textbf{70.0} & \textbf{62.7} & \textbf{54.6} & 49.7 & \textbf{55.6} & \textbf{74.0} & 51.1 & \textbf{37.5} & 63.4 & 1.2 & 69.9 & \textbf{68.5} & 58.0 & 70.9 & 48.8 & \textbf{73.9} & 19.4 & \textbf{42.9} & \textbf{60.8} & \textbf{54.5} & \textbf{+2.7}\\
\hline
Faster R-CNN~\cite{faster_rcnn}& large & 91.8 & 86.4 & 81.7 & 57.7 & 79.8 & 89.0 & 87.7 & \textbf{91.0} & 58.9 & 76.6 & 62.5 & \textbf{90.5} & 87.2 & 90.4 & 87.8 & 51.8 & 82.9 & 73.6 & 85.2 & 78.9 & 79.6 & 0.0\\
StuffNet-10 & large & 90.1 & \textbf{87.6} & 83.2 & \textbf{65.4} & 83.0 & 87.4 & \textbf{89.2} & 89.4 & \textbf{65.1} & \textbf{82.0} & 64.2 & 89.3 & 86.7 & \textbf{91.1} & 88.1 & \textbf{63.0} & 81.9 & 74.8 & \textbf{90.0} & \textbf{81.7} & \textbf{81.7} & \textbf{+2.1}\\
StuffNet-30 & large & \textbf{92.6} & 85.9 & \textbf{83.3} & 60.9 & \textbf{83.3} & \textbf{90.4} & 89.1 & 90.7 & 64.0 & 76.8 & \textbf{65.0} & 89.1 & \textbf{87.3} & 87.8 & \textbf{88.2} & 57.1 & \textbf{83.5} & \textbf{76.9} & 89.6 & 78.4 & 81.0 & 1.4\\
\hline
\end{tabular}}
\caption{Analyzing StuffNet performance gains for objects of different sizes. Trained on Pascal VOC 2010 \texttt{train}, tested on Pascal VOC 2010 \texttt{val}}
\label{tab:2}
\end{table*}
\begin{table*}[ht!]
\resizebox{\textwidth}{!}{
\begin{tabular}{|c|c|cccccccccccccccccccc|c|}
\hline
\textbf{Dataset} & \textbf{Method} & \textbf{areo} & \textbf{bike} & \textbf{bird} & \textbf{boat} & \textbf{bottle} & \textbf{bus} & \textbf{car} & \textbf{cat} & \textbf{chair} & \textbf{cow} & \textbf{table} & \textbf{dog} & \textbf{horse} & \textbf{mbike} & \textbf{person} & \textbf{plant} & \textbf{sheep} & \textbf{sofa} & \textbf{train} & \textbf{tv} & \textbf{mAP}\\
\hline
\multirow{7}{*}{VOC 07} & YOLO~\cite{yolo} & \multicolumn{20}{c}{--} & 63.4\\
& MR-CNN \& S-CNN~\cite{komodakis_context} & \textbf{76.8} & 75.7 & 67.6 & 55.1 & 45.6 & 77.6 & 76.5 & 78.4 & 46.7 & 74.7 & 68.8 & 79.3 & 74.2 & 77.0 & 62.5 & 37.4 & 64.3 & 63.8 & 74.0 & \textbf{74.7} & 67.5\\
& SSD300~\cite{ssd} & 73.4 & 77.5 & 64.1 & 59.0 & 38.9 & 75.2 & 80.8 & 78.5 & 46.0 & 67.8 & 69.2 & 76.6 & 82.1 & 77.0 & 72.5 & 41.2 & 64.2 & 69.1 & 78.0 & 68.5 & 68.0\\
& SSD512~\cite{ssd} & 75.1 & 81.4 & 69.8 & \textbf{60.8} & 46.3 & \textbf{82.6} & \textbf{84.7} & \textbf{84.1} & 48.5 & 75.0 & 67.4 & 82.3 & 83.9 & \textbf{79.4} & 76.6 & \textbf{44.9} & 69.9 & 69.1 & 78.1 & 71.8 & 71.6\\
& OHEM~\cite{ohem} & 71.2 & 78.3 & 69.2 & 57.9 & 46.5 & 81.8 & 79.1 & 83.2 & 47.9 & 76.2 & 68.9 & 83.2 & 80.8 & 75.8 & 72.7 & 39.9 & 67.5 & 66.2 & 75.6 & 75.9 & 69.9\\
& Faster R-CNN~\cite{faster_rcnn} & 70.0 & 80.6 & 70.1 & 57.3 & 49.9 & 78.2 & 80.4 & 82.0 & \textbf{52.2} & 75.3 & 67.2 & 80.3 & 79.8 & 75.0 & 76.3 & 39.1 & 68.3 & 67.3 & \textbf{81.1} & 67.6 & 69.9\\
\cline{2-23}
& StuffNet-30 & 72.6 & \textbf{81.7} & \textbf{70.6} & 60.5 & \textbf{53.0} & 81.5 & 83.7 & 83.9 & \textbf{52.2} & \textbf{78.9} & \textbf{70.7} & \textbf{85.0} & \textbf{85.7} & 77.0 & \textbf{78.7} & 42.2 & \textbf{73.6} & \textbf{69.2} & 79.2 & 73.8 & \textbf{72.7}\\ 
\hhline{|=|=|====================|=|}
\multirow{4}{*}{VOC 10} & segDeepM-16 layers~\cite{segdeepm} & 82.3 & 75.2 & 67.1 & 50.7 & 49.8 & 71.1 & 69.5 & \textbf{88.2} & 42.5 & 71.2 & 50.0 & 85.7 & 76.6 & 81.8 & 69.3 & 41.5 & 71.9 & \textbf{62.2} & 73.2 & \textbf{64.6} & 67.2\\
& Faster R-CNN~\cite{faster_rcnn} & 80.9 & 75.0 & 70.6 & 54.8 & 52.9 & 74.0 & 74.3 & 87.2 & 46.9 & \textbf{72.6} & 50.4 & 85.9 & 78.2 & 83.0 & 79.0 & 41.5 & \textbf{74.1} & 58.4 & 79.0 & 63.5 & 69.1\\
\cline{2-23}
& StuffNet-10 \href{http://host.robots.ox.ac.uk:8080/anonymous/V2VBFC.html}{$\dagger$} & 82.3 & 75.9 & \textbf{72.1} & \textbf{56.3} & 54.0 & \textbf{77.5} & \textbf{75.7} & 86.9 & 47.9 & 72.1 & 52.2 & \textbf{86.2} & 78.4 & 83.0 & \textbf{79.4} & \textbf{45.4} & 72.4 & 58.5 & 78.2 & 64.3 & 69.9\\
& StuffNet-30 \href{http://host.robots.ox.ac.uk:8080/anonymous/BVBKNF.html}{$\ddagger$} & \textbf{83.0} & \textbf{77.1} & 71.9 & 55.5 & \textbf{54.5} & 76.4 & 75.0 & 87.4 & \textbf{48.7} & 70.8 & \textbf{54.6} & 85.5 & \textbf{79.3} & \textbf{83.2} & 79.2 & 44.6 & 73.3 & 60.0 & \textbf{79.8} & 62.3 & \textbf{70.1}\\
\hhline{|=|=|====================|=|}
\multirow{3}{*}{VOC 12} & OHEM~\cite{ohem} & 81.5 & \textbf{78.9} & 69.6 & \textbf{52.3} & 46.5 & \textbf{77.4} & 72.1 & \textbf{88.2} & \textbf{48.8} & 73.8 & \textbf{58.3} & \textbf{86.9} & 79.7 & \textbf{81.4} & 75.0 & 43.0 & 69.5 & \textbf{64.8} & \textbf{78.5} & 68.9 & 69.8\\
& Faster R-CNN~\cite{faster_rcnn} & 82.3 & 76.4 & 71.0 & 48.4 & 45.2 & 72.1 & 72.3 & 87.3 & 42.2 & 73.7 & 50.0 & 86.8 & 78.7 & 78.4 & 77.4 & 34.5 & 70.1 & 57.1 & 77.1 & 58.9 & 67.0\\
\cline{2-23}
& StuffNet-30 \href{http://host.robots.ox.ac.uk:8080/anonymous/MN0IF0.html}{$\star$} & \textbf{83.0} & 76.9 & \textbf{71.2} & 51.6 & \textbf{50.1} & 76.4 & \textbf{75.7} & 87.8 & 48.3 & \textbf{74.8} & 55.7 & 85.7 & \textbf{81.2} & 80.3 & \textbf{79.5} & \textbf{44.2} & \textbf{71.8} & 61.0 & \textbf{78.5} & \textbf{65.4} & \textbf{70.0}\\
\hline
\end{tabular}}
\caption{StuffNet comparison with baselines on various datasets. For each dataset, the networks were trained on the respective \texttt{trainval} split and tested on the \texttt{test} split.}
\label{tab:3}
\end{table*}
\begin{table}[ht!]
\resizebox{0.49\textwidth}{!}{
\begin{tabular}{|c|cccccccccc|c|}
\hline
\textbf{Method} & \textbf{wall} & \textbf{floor} & \textbf{water} & \textbf{tree} & \textbf{sky} & \textbf{road} & \textbf{ground} & \textbf{building} & \textbf{mountain} & \textbf{mAP}\\
\hline
StuffNet-10 & 48.1 & \textbf{39.4} & 68.8 & 64.8 & 84.6 & \textbf{39.0} & 61.4 & 40.6 & 40.4 & 54.1\\
Separate & \textbf{48.6} &  38.9 &  \textbf{69.3} &  \textbf{64.9} & \textbf{85.1} & 37.8 & \textbf{62.3} & \textbf{41.3} & \textbf{40.6} & \textbf{54.3}\\
\hline
StuffNet-30 & 66.7 & 55.3 & 75.4 & 77.3 & 90.0 & 61.0 & 73.8 & 59.9 & 59.2 & 68.7\\
\hline
\end{tabular}}
\caption{Stuff segmentation performance results. \texttt{Separate} indicates a DeepLab network trained separately for segmenting the 10 stuff classes, using training parameters described in Section~\ref{sec:3-3-2}. \texttt{StuffNet-10} and \texttt{Separate} were trained on Pascal VOC 2010~\texttt{train} and tested on Pascal VOC 2010~\texttt{val}. \texttt{StuffNet-30} was trained on Pascal VOC 2012 \texttt{trainval} using the feature constraining method described in Section~\ref{sec:3-3-2} and tested on Pascal VOC 2012 \texttt{test}.}
\label{tab:4}
\end{table}
We implemented StuffNet by modifying the publicly available Python code from Faster R-CNN, which uses the Caffe library~\cite{caffe}. All our experiments use 70K training iterations. The learning rate starts at 1e-3 and drops by a factor of 10 after 50K iterations. We use a momentum of 0.9 and weight decay of 0.0005, as is standard practice. We initialize Faster R-CNN from the public VGG 16-layer model trained on ImageNet. Convolution layers specific to DeepLab are initialized using their public `init' model, which is modified from VGG-16. The RPN and final fully connected layers are initialized randomly using Xavier~\cite{xavier} initialization. In this section we evaluate the various StuffNet variants on the their object detection performance on a variety of datasets. For all our experiments, we use Faster R-CNN as a common baseline.
\subsection{Impact of multi-task learning and multi-feature classification}
StuffNet relies on 1) multi-task learning and 2) multi-feature classification to improve object detection performance. To analyse the impact of multi-task learning and multi-feature classification, we train the three StuffNet variants on Pascal VOC 2010 train split, and present evaluation results on the val split in Table~\ref{tab:1}. Multi-task learning (\texttt{StuffNet-multitask}) improves performance by 0.5\% mAP, while it's combination with multi-feature classification (\texttt{StuffNet-10}) improves performance by 1.5\% mAP. Adding in object segmentation features (\texttt{StuffNet-30}) delivers an additional 0.5\% mAP.
\subsection{Impact on different sized objects}
Cognitive psychology~\cite{olivia_context_survey} and computer vision~\cite{context_survey} research has shown that indentifying the surroundings is most helpful for identifying visually impoverished (i.e. blurred, partially occluded or small in size) objects. To analyze the impact of multi-task learning and multi-feature classification on the detection performance of visually impoverished objects, we evaluated Faster R-CNN, \texttt{StuffNet-10} and \texttt{StuffNet-30}, considering objects of three different sizes. As in the MS COCO~\cite{mscoco} dataset, the sizes for objects are defined as \texttt{small} (area $<$ 32x32 sq. pixels), \texttt{medium} (32x32 $<$ area $<$ 96x96 sq. pixels) and \texttt{large} (area $>$ 96x96 sq. pixels). When benchmarking on objects of each size, the ground truth labels for other sizes were marked `ignore' while evaluating each size using the metric presented in~\cite{dollar_eval}. Table~\ref{tab:2} shows that the increase in mAP by StuffNet against Faster R-CNN is highest for \texttt{small} objects (5.1 percentage points mAP increase). This shows empirically that multi-task learning and multi-feature classification improves the object detection performance especially for small-sized objects, which are usually visually impoverished and need extra contextual information to be identified.
\subsection{Evaluation on Datasets without stuff labels}
One of the attractive features of StuffNet is that once the segmentation branch has been trained, it can be used to train StuffNet on any dataset that does not have stuff labels. This is done using the feature constraining mechanism described in Section~\ref{sec:3-3-2}. In this section, we train \texttt{StuffNet-30} using this method on Pascal VOC 2007~\cite{voc-2007} and Pascal VOC 2012~\cite{voc-2012} \texttt{trainval} splits, and test them on the respective \texttt{test} splits. Results presented in Table~\ref{tab:3} show significant improvements delivered by StuffNet-30 compared to Faster R-CNN. We also compare the \texttt{StuffNet-30} performance against two recent baselines that use contextual information and diverse features for object detection: Multi-region and segmentation-aware CNN (MR-CNN \& S-CNN) from~\cite{komodakis_context} and segDeepM~\cite{segdeepm}. MR-CNN \& S-CNN combines features pooled from various contextual regions around the region proposal and features from a network trained \textit{separately} to segment the same objects that are being detected.~\cite{komodakis_context} also develops an iterative object re-localization strategy, that is applied as a post-processing step. Since object re-localization is beyond the scope of this paper, we compare StuffNet against the results from~\cite{komodakis_context} \textit{without} object re-localization. segDeepM diversifies the features input to the region classifier using features pooled from an expanded area around the region proposal, and extracted from the AlexNet~\cite{alexnet} CNN trained for the ImageNet image recognition challenge.\par
As shown in Table~\ref{tab:3}, compared to Faster R-CNN \texttt{StuffNet-30} achieves a \textit{performance gain of 2.8 percentage points on VOC 2007 and 3.0 percentage points on VOC 2012}, both of which have no stuff labels. In contrast, the comparable models from~\cite{komodakis_context} and~\cite{segdeepm} perform worse than Faster R-CNN by around 2\% mAP. This demonstrates that multi-task learning and constraining features to output segmentations learnt from VOC 2010 act as an effective regularizer for the the object detection task. Figure~\ref{fig:4} shows some examples of \texttt{StuffNet-30} detections on images from Pascal VOC 2010 \texttt{test}, compared to Faster R-CNN. Figure~\ref{fig:5} shows detection (Faster R-CNN and \texttt{StuffNet-30}) and segmentation (\texttt{StuffNet-30}) results on images from the MS COCO dataset, using models trained on Pascal VOC 2010 \texttt{trainval}. Note that \texttt{StuffNet-30} is particularly better than Faster R-CNN at detecting small objects.   
\subsection{Semantic segmentation evaluation}
Table~\ref{tab:4} shows the performance of \texttt{StuffNet-10} and a separately trained DeepLab network on the task of segmenting 10 stuff classes. StuffNet performance is lower than separately trained DeepLab by 0.2 percentage points. We hypothesize that an appropriately weighted sum of the Faster R-CNN and DeepLab cost functions while training StuffNet will improve semantic segmentation performance. This is because both losses are normalized by the number of datapoints contributing to the sum (number of region proposals in mini-batch for Faster R-CNN and number of pixels in the image for DeepLab). This normalization weighs the loss for a single pixel's wrong semantic segmentation output with a very low factor. A careful search for the relative weights of the Faster R-CNN and DeepLab loss functions can be done using cross validation.\par
\begin{figure*}[h!]
  \centering
  \begin{subfigure}[b]{0.24\textwidth}
    \includegraphics[width=\textwidth]{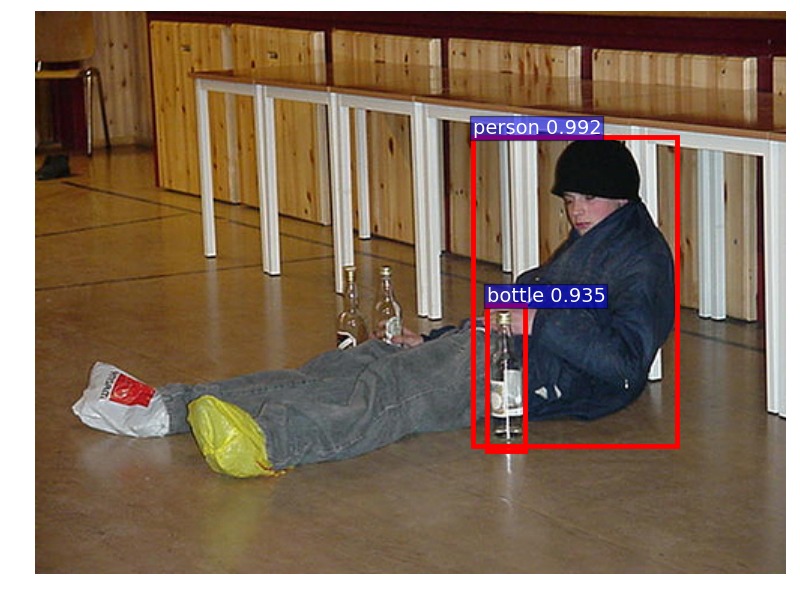}
  \end{subfigure}
  \begin{subfigure}[b]{0.24\textwidth}
    \includegraphics[width=\textwidth]{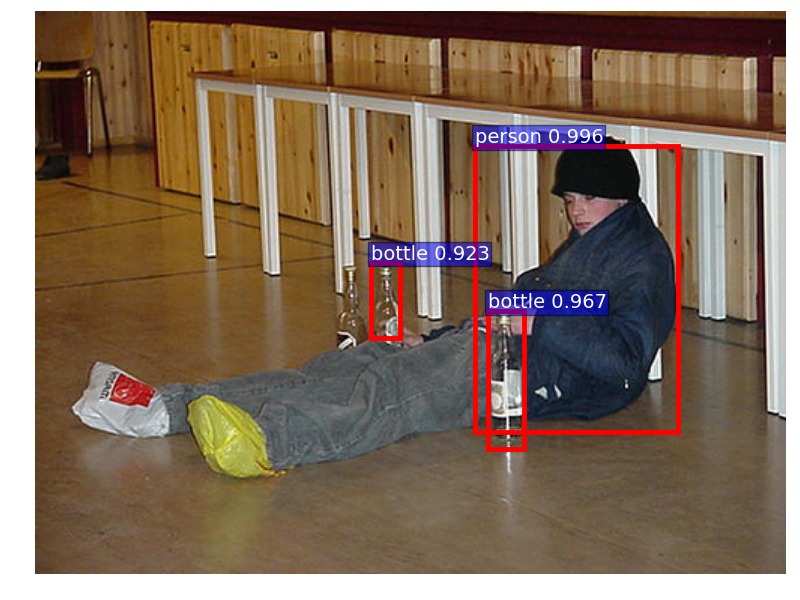}
  \end{subfigure}
  \begin{subfigure}[b]{0.24\textwidth}
    \includegraphics[width=\textwidth]{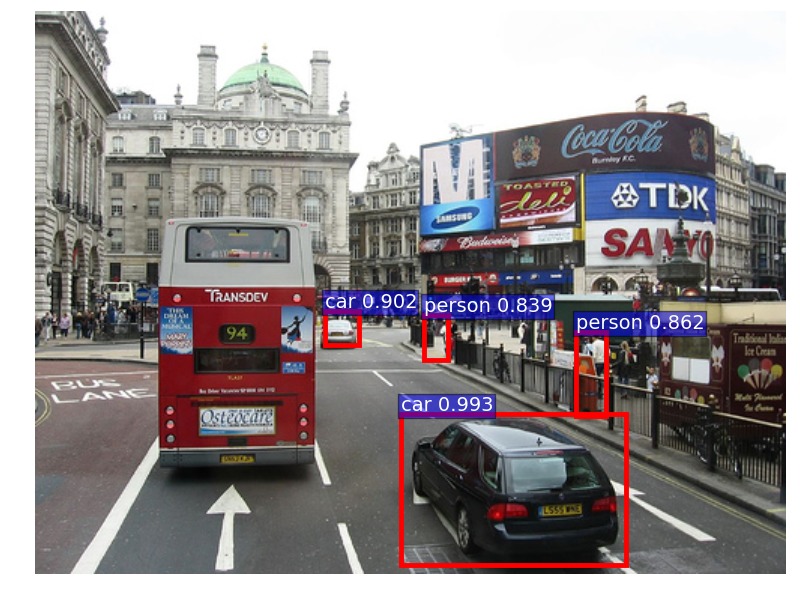}
  \end{subfigure}
  \begin{subfigure}[b]{0.24\textwidth}
    \includegraphics[width=\textwidth]{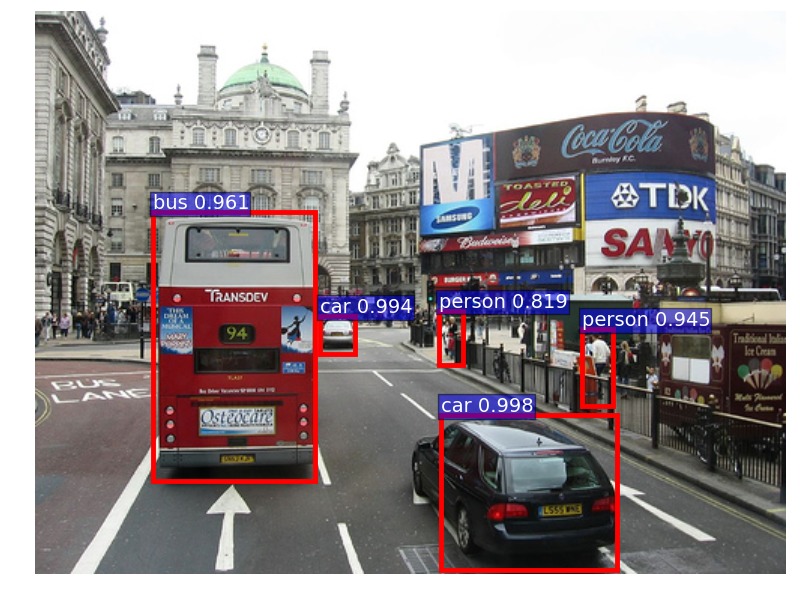}
  \end{subfigure}
  \begin{subfigure}[b]{0.24\textwidth}
    \includegraphics[width=\textwidth]{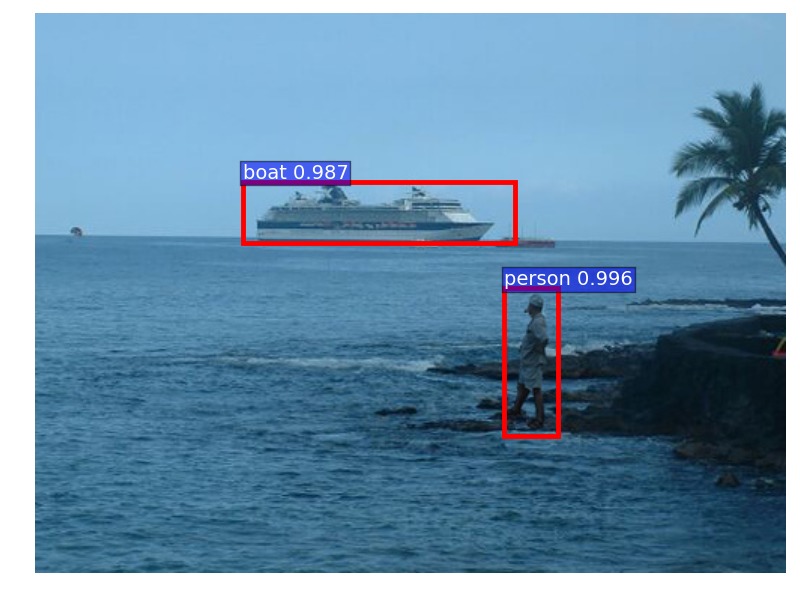}
  \end{subfigure}
  \begin{subfigure}[b]{0.24\textwidth}
    \includegraphics[width=\textwidth]{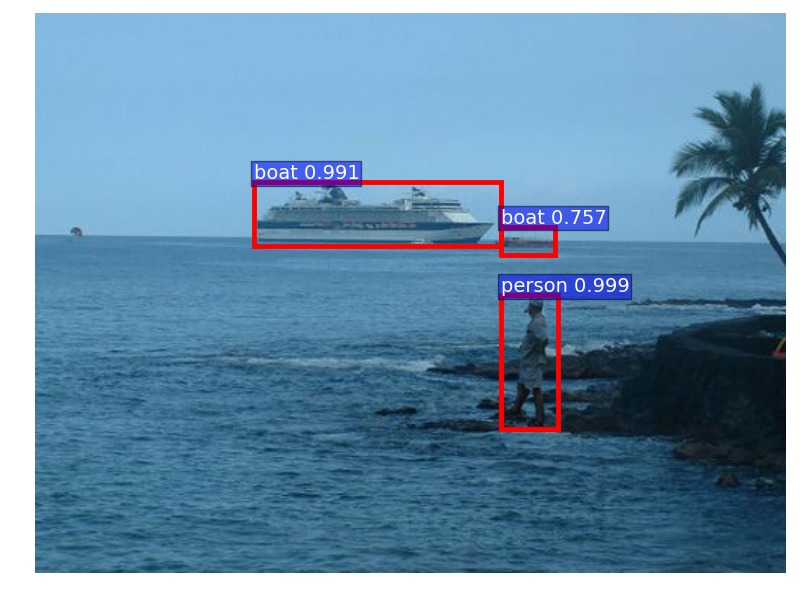}
  \end{subfigure}
  \begin{subfigure}[b]{0.24\textwidth}
    \includegraphics[width=\textwidth]{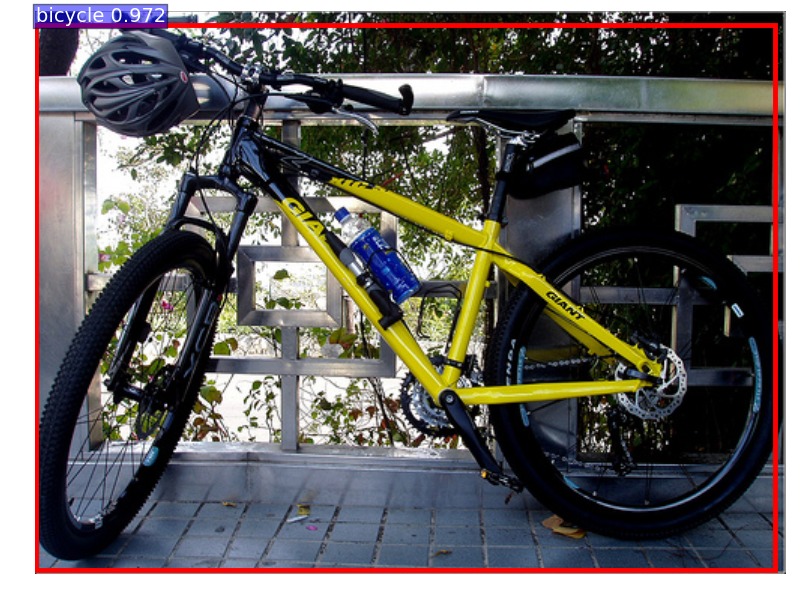}
  \end{subfigure}
  \begin{subfigure}[b]{0.24\textwidth}
    \includegraphics[width=\textwidth]{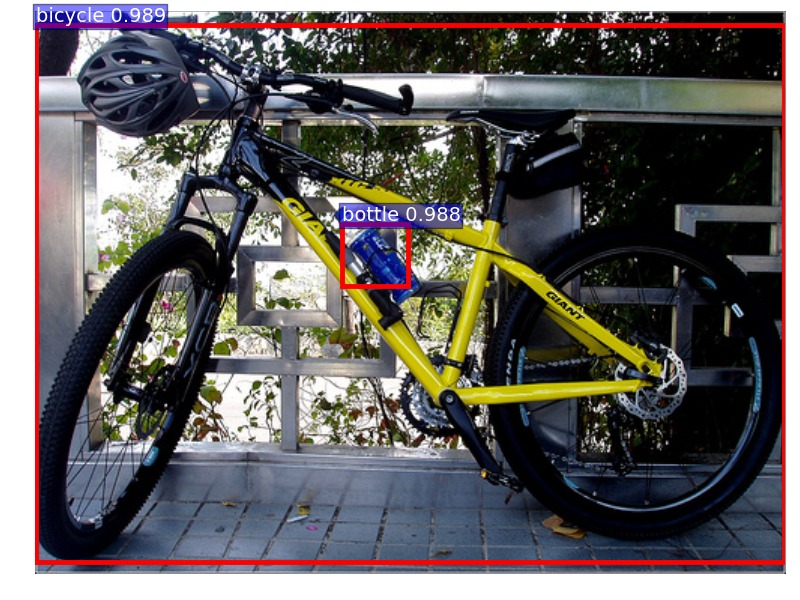}
  \end{subfigure}
  \begin{subfigure}[b]{0.24\textwidth}
    \includegraphics[width=\textwidth]{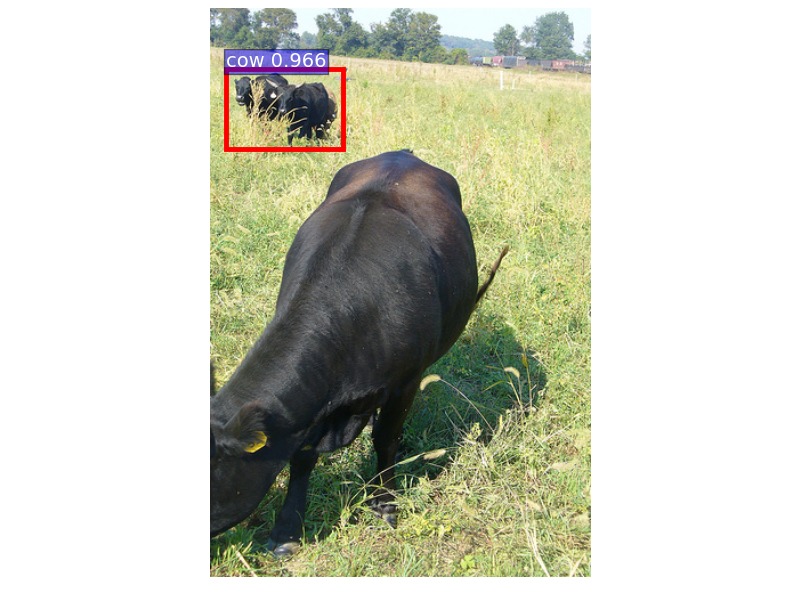}
  \end{subfigure}
  \begin{subfigure}[b]{0.24\textwidth}
    \includegraphics[width=\textwidth]{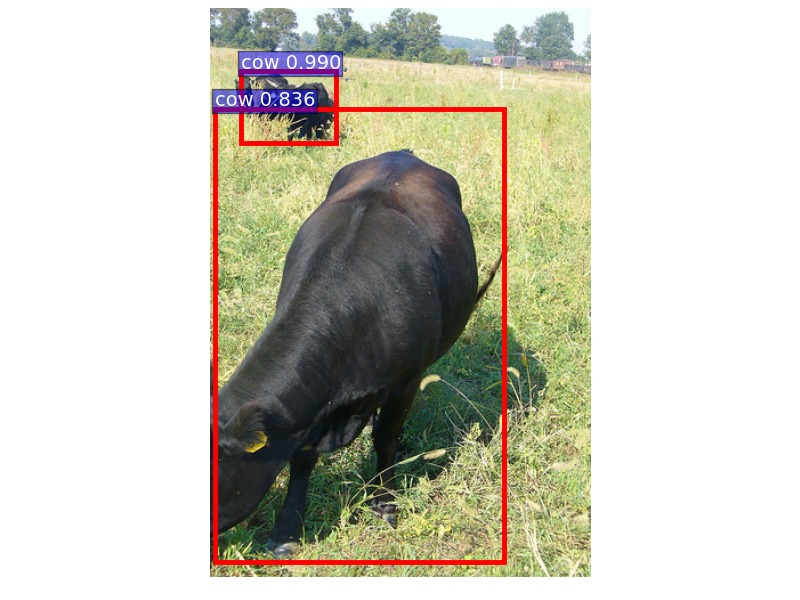}
  \end{subfigure}  
  \begin{subfigure}[b]{0.24\textwidth}
    \includegraphics[width=\textwidth]{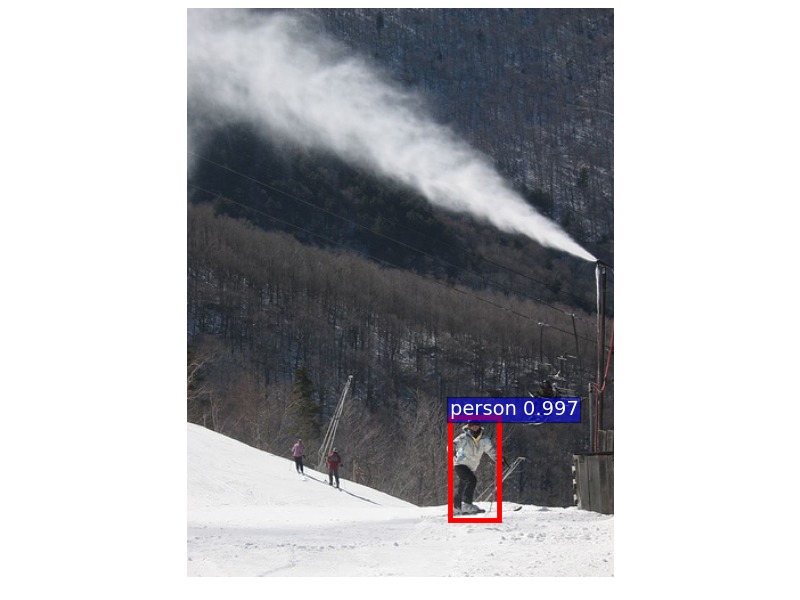}
  \end{subfigure}
  \begin{subfigure}[b]{0.24\textwidth}
    \includegraphics[width=\textwidth]{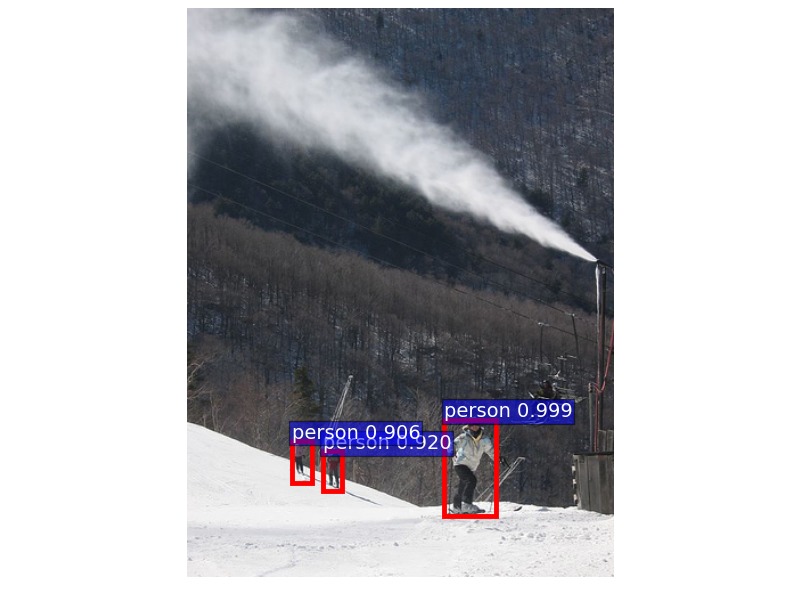}
  \end{subfigure}
\caption{Pairs of Faster R-CNN (left) vs. \texttt{StuffNet-30} (right) detections on sample images from the Pascal VOC 2010 \texttt{test}. Both models have been trained on Pascal VOC 2010 \texttt{trainval}}
\label{fig:4}
\end{figure*}

The last line of Table~\ref{tab:4} shows the performance of a \texttt{StuffNet-30} trained with the feature constraining method on Pascal VOC 2012, a dataset that does not have stuff labels. The ground-truth labels for this evaluation were generated using \texttt{StuffNet-30} trained on Pascal VOC 2010 \texttt{trainval}. The high performance demonstrates that the feature constraining method is effective. Figure~\ref{fig:4} shows some examples of object and stuff semantic segmentation using StuffNet models trained both normally and using feature constraining.
\FloatBarrier
\section{Conclusion} \label{sec:5}
We presented StuffNet, an algorithm that uses multi-task learning and multi-feature classification to improve object detection, especially for small objects. StuffNet allows stuff and object semantic segmentation to closely influence the object detection process. The applicability of StuffNet can be extended to datasets that lack stuff or object segmentation labels by using a simple feature constraining procedure.\par
This paper opens up a few avenues for future research: 1) in the current architecture, the RPN does not benefit from multi-feature inputs. We conjecture that allowing this will improve localization performance, especially for small objects. 2) using deeper representations (e.g. ResNet~\cite{resnet}) should also further improve performance.
\begin{figure*}[h!]
  \centering
	\begin{subfigure}[b]{0.3\textwidth}
    	\includegraphics[width=\textwidth]{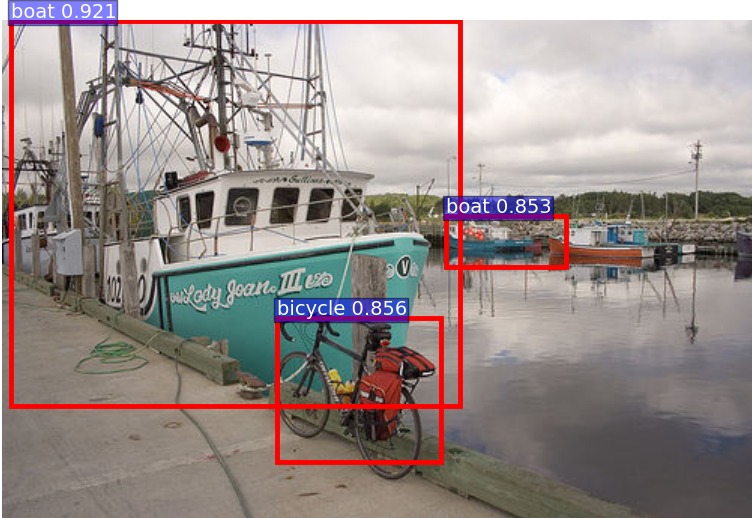}
    \includegraphics[width=\textwidth]{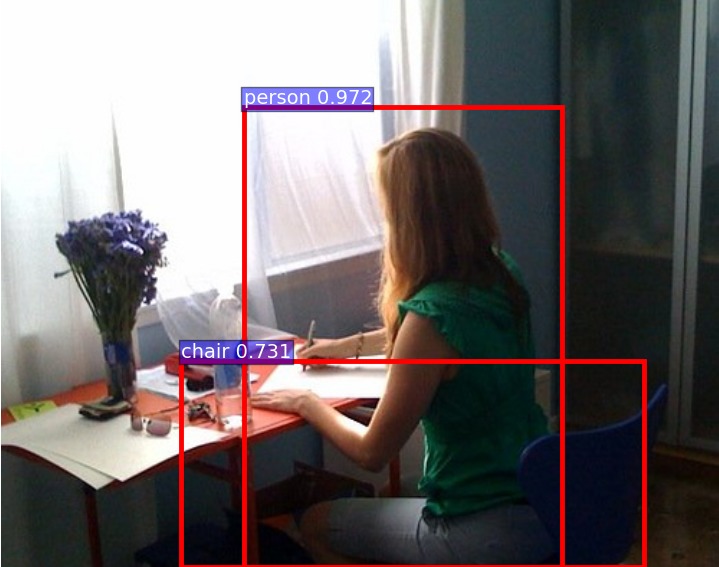}
    \includegraphics[width=\textwidth]{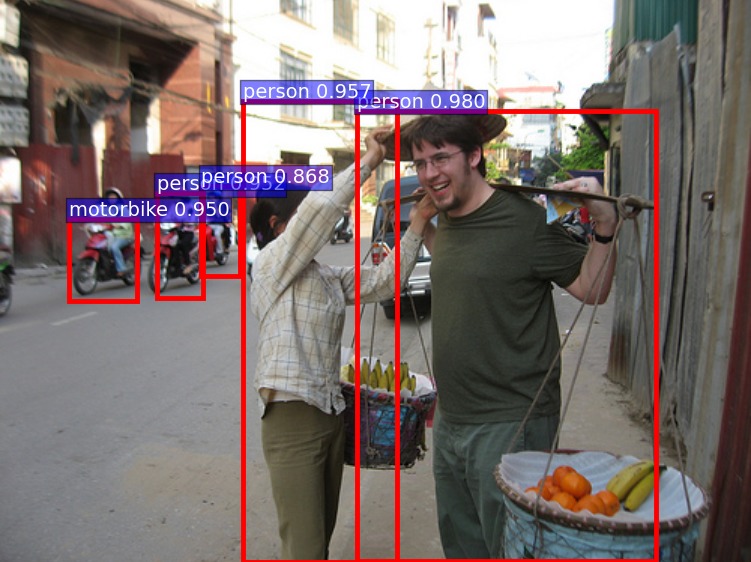}
    \includegraphics[width=\textwidth]{images/supp/3_f}
    \includegraphics[width=\textwidth]{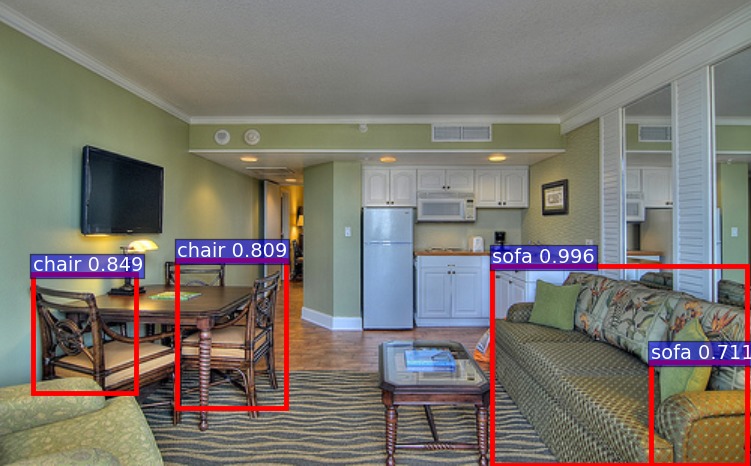}
    \includegraphics[width=\textwidth]{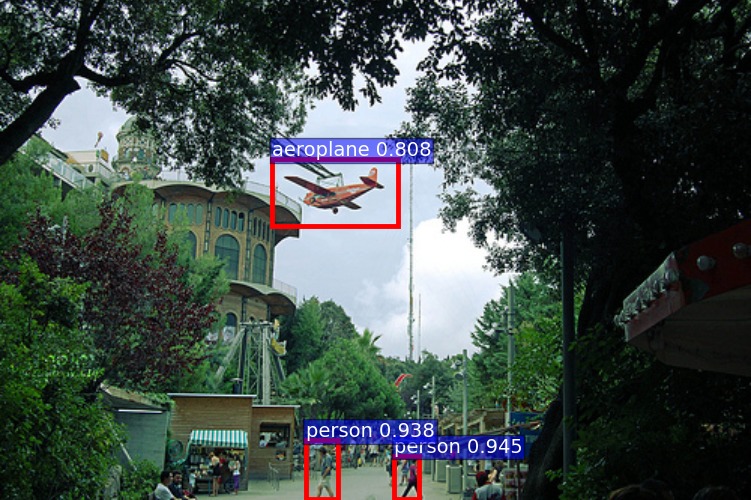}
    \caption{Faster R-CNN}
	\end{subfigure}
	\begin{subfigure}[b]{0.3\textwidth}
    	\includegraphics[width=\textwidth]{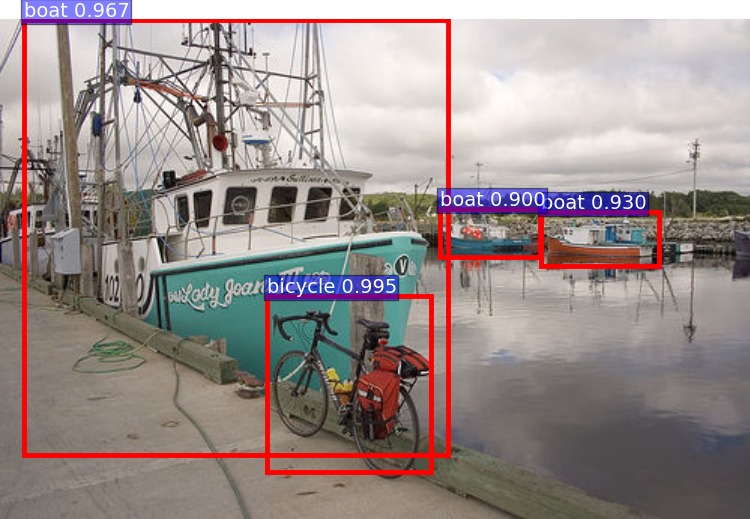}
    \includegraphics[width=\textwidth]{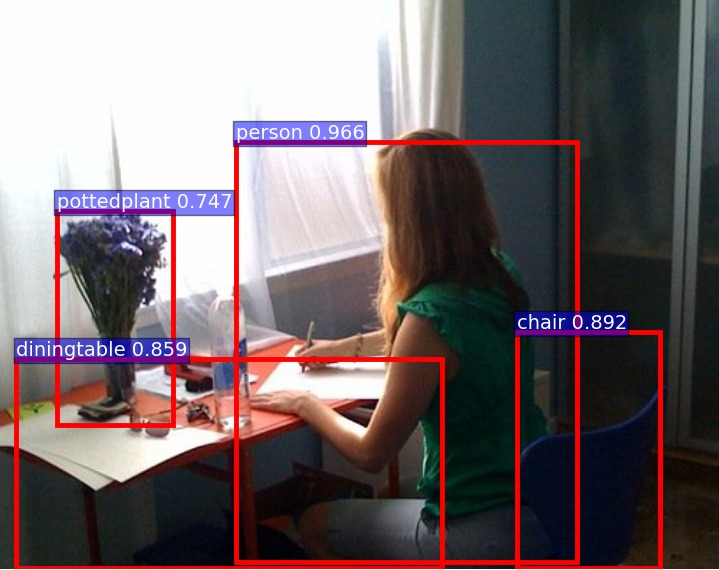}
    \includegraphics[width=\textwidth]{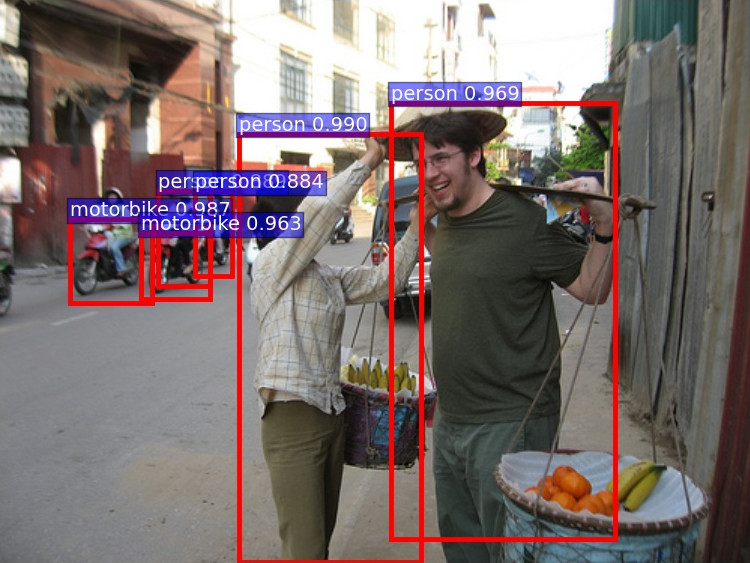}
    \includegraphics[width=\textwidth]{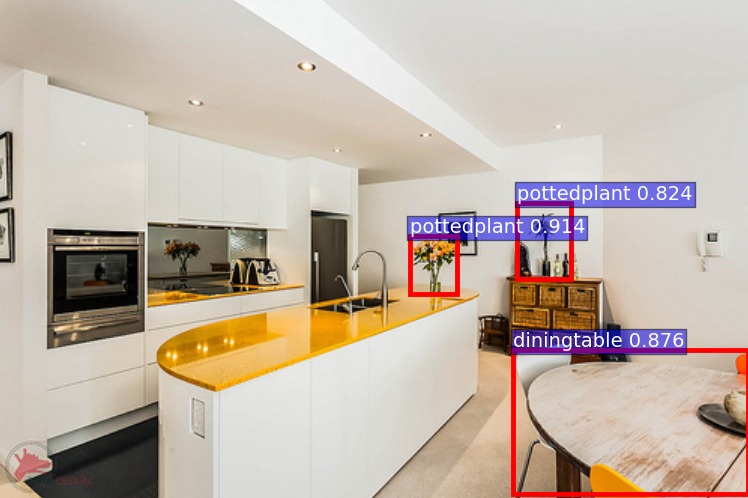}
    \includegraphics[width=\textwidth]{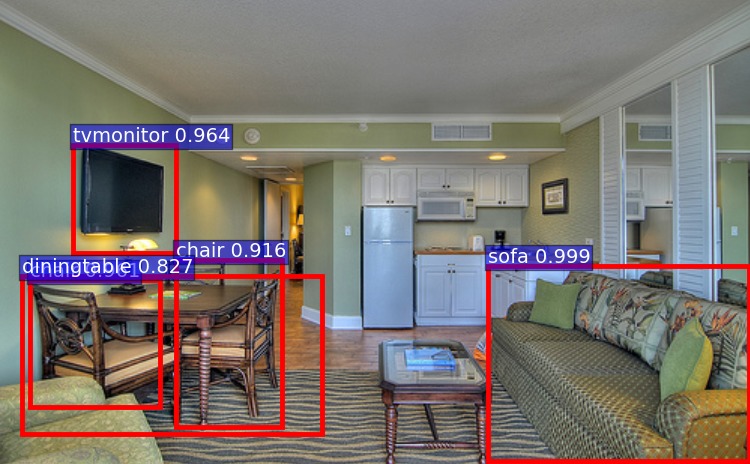}
    \includegraphics[width=\textwidth]{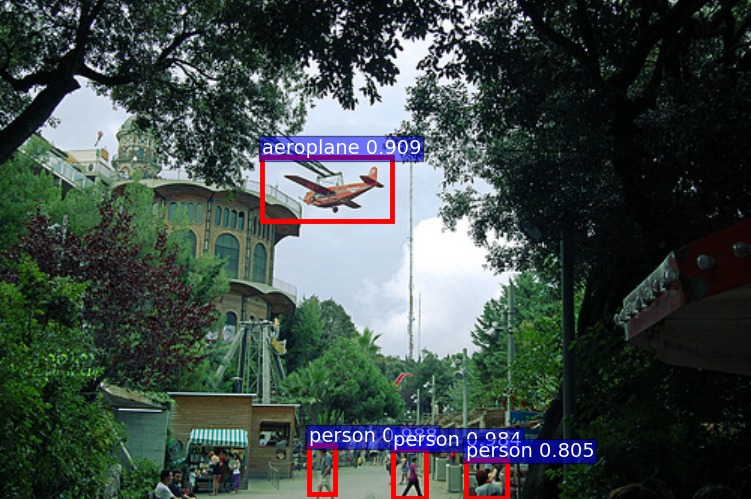}
    \caption{\texttt{StuffNet-30}}
	\end{subfigure}	
	\begin{subfigure}[b]{0.3\textwidth}
    	\includegraphics[width=\textwidth]{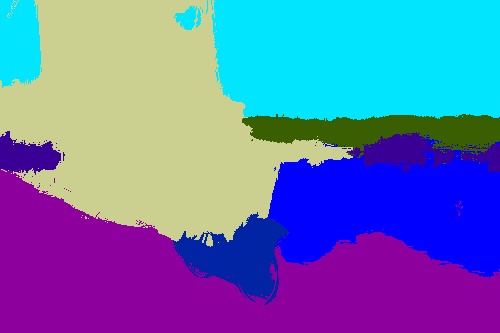}
    \includegraphics[width=\textwidth]{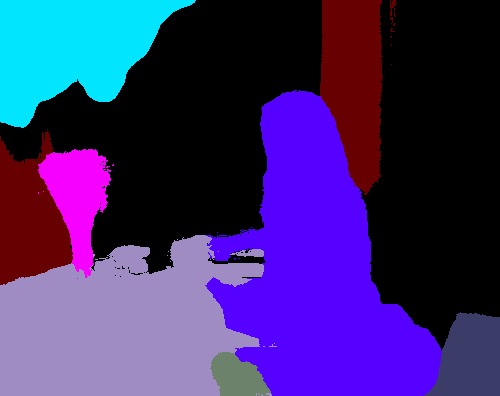}
    \includegraphics[width=\textwidth]{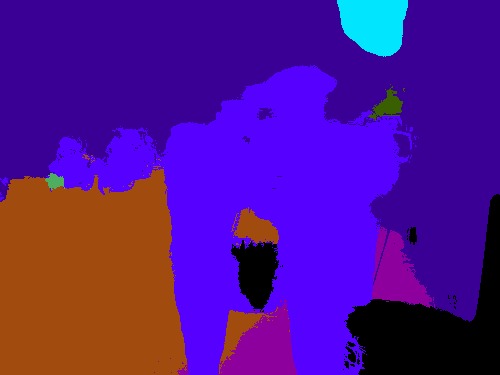}
    \includegraphics[width=\textwidth]{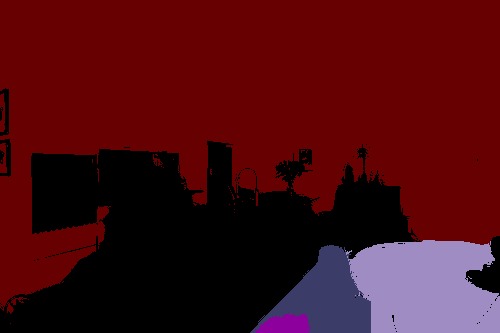}
    \includegraphics[width=\textwidth]{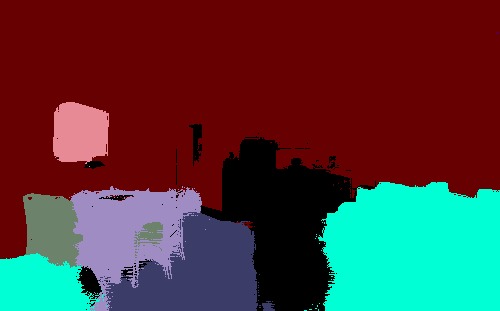}
    \includegraphics[width=\textwidth]{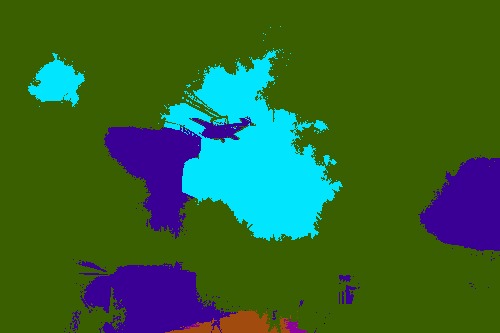}
    \caption{~\texttt{StuffNet-30}}
	\end{subfigure}	
	\begin{subfigure}[b]{0.08\textwidth}
			\includegraphics[width=\textwidth]{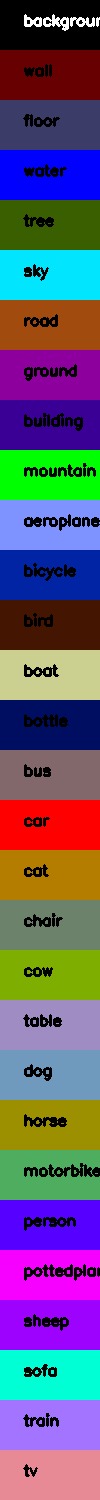}
			\caption{Legend}
	\end{subfigure}	
\caption{\texttt{StuffNet-30} detection and segmentation qualitative results on images from the MS COCO dataset. The model was trained on Pascal VOC 2010 \texttt{trainval}. We also show Faster R-CNN detections in the first column for comparison. \texttt{StuffNet-30} performs particularly well in detecting small objects.}
\label{fig:6}
\end{figure*}

\clearpage

{\small
\bibliographystyle{templates/ieee}
\bibliography{references}

\begin{thebibliography}{10}\itemsep=-1pt

\bibitem{cityscapes}
M.~Cordts, M.~Omran, S.~Ramos, T.~Rehfeld, M.~Enzweiler, R.~Benenson,
  U.~Franke, S.~Roth, and B.~Schiele.
\newblock The cityscapes dataset for semantic urban scene understanding.
\newblock In {\em Proc. of the IEEE Conference on Computer Vision and Pattern
  Recognition (CVPR)}, 2016.

\bibitem{resnet_seg}
J.~Dai, K.~He, and J.~Sun.
\newblock Instance-aware semantic segmentation via multi-task network cascades.
\newblock {\em arXiv preprint arXiv:1512.04412}, 2015.

\bibitem{context_survey}
S.~K. Divvala, D.~Hoiem, J.~H. Hays, A.~A. Efros, and M.~Hebert.
\newblock An empirical study of context in object detection.
\newblock In {\em Computer Vision and Pattern Recognition, 2009. CVPR 2009.
  IEEE Conference on}, pages 1271--1278. IEEE, 2009.

\bibitem{dollar_eval}
P.~Dollar, C.~Wojek, B.~Schiele, and P.~Perona.
\newblock Pedestrian detection: An evaluation of the state of the art.
\newblock {\em IEEE transactions on pattern analysis and machine intelligence},
  34(4):743--761, 2012.

\bibitem{voc-2007}
M.~Everingham, L.~Van~Gool, C.~K.~I. Williams, J.~Winn, and A.~Zisserman.
\newblock The {PASCAL} {V}isual {O}bject {C}lasses {C}hallenge 2007 {(VOC2007)}
  {R}esults.
\newblock
  http://www.pascal-network.org/challenges/VOC/voc2007/workshop/index.html.

\bibitem{voc-2010}
M.~Everingham, L.~Van~Gool, C.~K.~I. Williams, J.~Winn, and A.~Zisserman.
\newblock The {PASCAL} {V}isual {O}bject {C}lasses {C}hallenge 2010 {(VOC2010)}
  {R}esults.
\newblock
  http://www.pascal-network.org/challenges/VOC/voc2010/workshop/index.html.

\bibitem{voc-2012}
M.~Everingham, L.~Van~Gool, C.~K.~I. Williams, J.~Winn, and A.~Zisserman.
\newblock The {PASCAL} {V}isual {O}bject {C}lasses {C}hallenge 2012 {(VOC2012)}
  {R}esults.
\newblock
  http://www.pascal-network.org/challenges/VOC/voc2012/workshop/index.html.

\bibitem{farhadi_attributes}
A.~Farhadi, I.~Endres, D.~Hoiem, and D.~Forsyth.
\newblock Describing objects by their attributes.
\newblock In {\em Computer Vision and Pattern Recognition, 2009. CVPR 2009.
  IEEE Conference on}, pages 1778--1785. IEEE, 2009.

\bibitem{komodakis_context}
S.~Gidaris and N.~Komodakis.
\newblock Object detection via a multi-region and semantic segmentation-aware
  cnn model.
\newblock In {\em Proceedings of the IEEE International Conference on Computer
  Vision}, pages 1134--1142, 2015.

\bibitem{fast_rcnn}
R.~Girshick.
\newblock Fast r-cnn.
\newblock In {\em Proceedings of the IEEE International Conference on Computer
  Vision}, pages 1440--1448, 2015.

\bibitem{xavier}
X.~Glorot and Y.~Bengio.
\newblock Understanding the difficulty of training deep feedforward neural
  networks.
\newblock In {\em International Conference on Artificial Intelligence and
  Statistics}, pages 249--256, 2010.

\bibitem{malik_small_context}
S.~Gupta, B.~Hariharan, and J.~Malik.
\newblock Exploring person context and local scene context for object
  detection.
\newblock {\em CoRR}, abs/1511.08177, 2015.

\bibitem{resnet}
K.~He, X.~Zhang, S.~Ren, and J.~Sun.
\newblock Deep residual learning for image recognition.
\newblock {\em arXiv preprint arXiv:1512.03385}, 2015.

\bibitem{sppnet}
K.~He, X.~Zhang, S.~Ren, and J.~Sun.
\newblock Spatial pyramid pooling in deep convolutional networks for visual
  recognition.
\newblock {\em Pattern Analysis and Machine Intelligence, IEEE Transactions
  on}, 37(9):1904--1916, 2015.

\bibitem{stuff}
G.~Heitz and D.~Koller.
\newblock Learning spatial context: Using stuff to find things.
\newblock In {\em European conference on computer vision}, pages 30--43.
  Springer, 2008.

\bibitem{caffe}
Y.~Jia, E.~Shelhamer, J.~Donahue, S.~Karayev, J.~Long, R.~Girshick,
  S.~Guadarrama, and T.~Darrell.
\newblock Caffe: Convolutional architecture for fast feature embedding.
\newblock {\em arXiv preprint arXiv:1408.5093}, 2014.

\bibitem{multi_3}
Z.~Kang, K.~Grauman, and F.~Sha.
\newblock Learning with whom to share in multi-task feature learning.
\newblock In {\em Proceedings of the 28th International Conference on Machine
  Learning (ICML-11)}, pages 521--528, 2011.

\bibitem{dense_crf}
P.~Kr{\"a}henb{\"u}hl and V.~Koltun.
\newblock Parameter learning and convergent inference for dense random fields.

\bibitem{imagenet}
A.~Krizhevsky, I.~Sutskever, and G.~E. Hinton.
\newblock Imagenet classification with deep convolutional neural networks.
\newblock In {\em Advances in neural information processing systems}, pages
  1097--1105, 2012.

\bibitem{alexnet}
A.~Krizhevsky, I.~Sutskever, and G.~E. Hinton.
\newblock Imagenet classification with deep convolutional neural networks.
\newblock In {\em Advances in neural information processing systems}, pages
  1097--1105, 2012.

\bibitem{deeplab}
C.~Liang-Chieh, G.~Papandreou, I.~Kokkinos, K.~Murphy, and A.~Yuille.
\newblock Semantic image segmentation with deep convolutional nets and fully
  connected crfs.
\newblock In {\em International Conference on Learning Representations}, 2015.

\bibitem{mscoco}
T.-Y. Lin, M.~Maire, S.~Belongie, J.~Hays, P.~Perona, D.~Ramanan,
  P.~Doll{\'a}r, and C.~L. Zitnick.
\newblock Microsoft coco: Common objects in context.
\newblock In {\em European Conference on Computer Vision}, pages 740--755.
  Springer, 2014.

\bibitem{ssd}
W.~Liu, D.~Anguelov, D.~Erhan, C.~Szegedy, S.~Reed, C.-Y. Fu, and A.~C. Berg.
\newblock {\em SSD: Single Shot MultiBox Detector}.
\newblock Springer International Publishing, 2016.

\bibitem{parsenet}
W.~Liu, A.~Rabinovich, and A.~C. Berg.
\newblock Parsenet: Looking wider to see better.
\newblock {\em arXiv preprint arXiv:1506.04579}, 2015.

\bibitem{fcn}
J.~Long, E.~Shelhamer, and T.~Darrell.
\newblock Fully convolutional networks for semantic segmentation.
\newblock In {\em Proceedings of the IEEE Conference on Computer Vision and
  Pattern Recognition}, pages 3431--3440, 2015.

\bibitem{holes}
S.~Mallat.
\newblock {\em A wavelet tour of signal processing: the sparse way}.
\newblock Academic press, 2008.

\bibitem{cross_stitch}
I.~Misra, A.~Shrivastava, A.~Gupta, and M.~Hebert.
\newblock Cross-stitch networks for multi-task learning.
\newblock {\em arXiv preprint arXiv:1604.03539}, 2016.

\bibitem{mottaghi}
R.~Mottaghi, X.~Chen, X.~Liu, N.-G. Cho, S.-W. Lee, S.~Fidler, R.~Urtasun, and
  A.~Yuille.
\newblock The role of context for object detection and semantic segmentation in
  the wild.
\newblock In {\em Proceedings of the IEEE Conference on Computer Vision and
  Pattern Recognition}, pages 891--898, 2014.

\bibitem{olivia_context_survey}
A.~Oliva and A.~Torralba.
\newblock The role of context in object recognition.
\newblock {\em Trends in cognitive sciences}, 11(12):520--527, 2007.

\bibitem{multi_1}
A.~Pentina, V.~Sharmanska, and C.~H. Lampert.
\newblock Curriculum learning of multiple tasks.
\newblock In {\em Proceedings of the IEEE Conference on Computer Vision and
  Pattern Recognition}, pages 5492--5500, 2015.

\bibitem{multi_4}
A.~Quattoni, M.~Collins, and T.~Darrell.
\newblock Transfer learning for image classification with sparse prototype
  representations.
\newblock In {\em Computer Vision and Pattern Recognition, 2008. CVPR 2008.
  IEEE Conference on}, pages 1--8. IEEE, 2008.

\bibitem{yolo}
J.~Redmon, S.~Divvala, R.~Girshick, and A.~Farhadi.
\newblock You only look once: Unified, real-time object detection.
\newblock In {\em The IEEE Conference on Computer Vision and Pattern
  Recognition (CVPR)}, June 2016.

\bibitem{faster_rcnn}
S.~Ren, K.~He, R.~Girshick, and J.~Sun.
\newblock Faster {R}-{CNN}: Towards real-time object detection with region
  proposal networks.
\newblock In {\em Advances in neural information processing systems}, pages
  91--99, 2015.

\bibitem{ohem}
A.~Shrivastava, A.~Gupta, and R.~Girshick.
\newblock Training region-based object detectors with online hard example
  mining.
\newblock In {\em The IEEE Conference on Computer Vision and Pattern
  Recognition (CVPR)}, June 2016.

\bibitem{vgg}
K.~Simonyan and A.~Zisserman.
\newblock Very deep convolutional networks for large-scale image recognition.
\newblock {\em CoRR}, abs/1409.1556, 2014.

\bibitem{selective_search}
J.~R. Uijlings, K.~E. van~de Sande, T.~Gevers, and A.~W. Smeulders.
\newblock Selective search for object recognition.
\newblock {\em International journal of computer vision}, 104(2):154--171,
  2013.

\bibitem{multi_2}
J.~Yim, H.~Jung, B.~Yoo, C.~Choi, D.~Park, and J.~Kim.
\newblock Rotating your face using multi-task deep neural network.
\newblock In {\em Proceedings of the IEEE Conference on Computer Vision and
  Pattern Recognition}, pages 676--684, 2015.

\bibitem{face2}
C.~Zhang and Z.~Zhang.
\newblock Improving multiview face detection with multi-task deep convolutional
  neural networks.
\newblock In {\em IEEE Winter Conference on Applications of Computer Vision},
  pages 1036--1041. IEEE, 2014.

\bibitem{face1}
Z.~Zhang, P.~Luo, C.~C. Loy, and X.~Tang.
\newblock Facial landmark detection by deep multi-task learning.
\newblock In {\em European Conference on Computer Vision}, pages 94--108.
  Springer, 2014.

\bibitem{cnn_crf}
S.~Zheng, S.~Jayasumana, B.~Romera-Paredes, V.~Vineet, Z.~Su, D.~Du, C.~Huang,
  and P.~H. Torr.
\newblock Conditional random fields as recurrent neural networks.
\newblock In {\em Proceedings of the IEEE International Conference on Computer
  Vision}, pages 1529--1537, 2015.

\bibitem{segdeepm}
Y.~Zhu, R.~Urtasun, R.~Salakhutdinov, and S.~Fidler.
\newblock segdeepm: Exploiting segmentation and context in deep neural networks
  for object detection.
\newblock In {\em Proceedings of the IEEE Conference on Computer Vision and
  Pattern Recognition}, pages 4703--4711, 2015.

\end{thebibliography}
}

\end{document}